\documentclass[11pt]{article}

\usepackage[final]{acl}
\usepackage{amsmath}

\usepackage{times}
\usepackage{latexsym}
\usepackage{listings}
\usepackage{xcolor}
\definecolor{codebg}{RGB}{248,248,248}
\usepackage{enumitem}   
\usepackage{graphicx}   
\usepackage{amssymb}  
\usepackage{algorithm}
\usepackage{algorithmic}
\usepackage{tabularx}
\usepackage{csvsimple}

\usepackage[T1]{fontenc}

\usepackage[utf8]{inputenc}

\usepackage{microtype}

\usepackage{inconsolata}
\usepackage{booktabs}   
\usepackage{multirow}   
\usepackage{graphicx}   

\usepackage{graphicx}
\usepackage{caption}
\usepackage{float}
\usepackage{booktabs}
\usepackage{enumitem}
\usepackage{amsmath}
\usepackage{amssymb}
\usepackage{svg}
\usepackage{listings}
\usepackage{xcolor}  

\definecolor{jsonkey}{rgb}{0.8, 0.1, 0.1}
\definecolor{jsonstring}{rgb}{0.1, 0.3, 0.7}
\definecolor{jsonnumber}{rgb}{0.1, 0.5, 0.1}
\definecolor{jsonnull}{rgb}{0.6, 0.0, 0.6}
\definecolor{jsonbg}{rgb}{0.98, 0.98, 0.97}

\lstdefinelanguage{json}{
    basicstyle=\ttfamily\small,
    backgroundcolor=\color{jsonbg},
    numbers=left,
    numberstyle=\scriptsize\color{gray},
    string=[s]{"}{"},
    showstringspaces=false,
    breaklines=true,
    frame=single,
    literate=
     *{0}{{{\color{jsonnumber}0}}}{1}
      {1}{{{\color{jsonnumber}1}}}{1}
      {2}{{{\color{jsonnumber}2}}}{1}
      {3}{{{\color{jsonnumber}3}}}{1}
      {4}{{{\color{jsonnumber}4}}}{1}
      {5}{{{\color{jsonnumber}5}}}{1}
      {6}{{{\color{jsonnumber}6}}}{1}
      {7}{{{\color{jsonnumber}7}}}{1}
      {8}{{{\color{jsonnumber}8}}}{1}
      {9}{{{\color{jsonnumber}9}}}{1}
      {:}{{{\color{black}{:}}}}{1}
      {,}{{{\color{black}{,}}}}{1}
      {\{}{{{\color{black}{\{}}}}{1}
      {\}}{{{\color{black}{\}}}}}{1}
      {[}{{{\color{black}{[}}}}{1}
      {]}{{{\color{black}{]}}}}{1}
      {"description"}{{{\color{jsonkey}"description"}}}{12}
      {"title"}{{{\color{jsonkey}"title"}}}{7}
      {"type"}{{{\color{jsonkey}"type"}}}{6}
      {"properties"}{{{\color{jsonkey}"properties"}}}{12}
      {"anyOf"}{{{\color{jsonkey}"anyOf"}}}{7}
      {"const"}{{{\color{jsonkey}"const"}}}{7}
      {"default"}{{{\color{jsonkey}"default"}}}{9}
      {"format"}{{{\color{jsonkey}"format"}}}{8}
      {"Market Value"}{{{\color{jsonstring}"Market Value"}}}{14}
      {"Currency Forward"}{{{\color{jsonstring}"Currency Forward"}}}{18}
      {"Currency Pair"}{{{\color{jsonstring}"Currency Pair"}}}{13}
      {null}{{{\color{jsonnull}null}}}{4}
}

\title{TASER: Table Agents for Schema‑guided Extraction and Recommendation}

\author{Nicole Cho \qquad Kirsty Fielding \qquad William Watson \\ {\bf Sumitra Ganesh} \qquad {\bf Manuela Veloso} \\
        J.P. Morgan AI Research \\
        \texttt{nicole.cho@jpmorgan.com}
}

\begin{document}
\maketitle
\begin{abstract}
Real-world financial filings report critical information about an entity's investment holdings, essential for assessing that entity's risk, profitability, and relationship profile.
Yet, these details are often buried in messy, multi-page, fragmented tables that are difficult to parse, hindering downstream QA and data normalization.
Specifically, 99.4\% of the tables in our financial table dataset lack bounding boxes, with the largest table 
spanning 44 pages. 
To address this, we present \textbf{TASER (Table Agents for Schema-guided Extraction and Recommendation)}, a continuously learning, agentic table extraction system that converts highly unstructured, multi-page, heterogeneous tables into normalized, schema-conforming outputs. 
Guided by an initial portfolio schema, TASER executes table detection, classification, extraction, and recommendations in a single pipeline.
Our Recommender Agent reviews unmatched outputs and proposes schema revisions, enabling TASER to outperform vision-based table detection models such as Table Transformer by 10.1\%. 
Within this continuous learning process, larger batch sizes yield a 104.3\% increase in useful schema recommendations and a 9.8\% increase in total extractions. 
To train TASER, we manually labeled 22,584 pages and 
3,213 tables covering \$731.7 billion in holdings, culminating in \textbf{TASERTab} to facilitate research on real-world financial tables and structured outputs. 
Our results highlight the promise of continuously learning agents for robust extractions from complex tabular data.
\end{abstract}

\section{Introduction}

\begin{figure}[t]
\centering
\includegraphics[width=0.87\columnwidth, trim=0.5cm 4cm 12cm 0.3cm, clip]{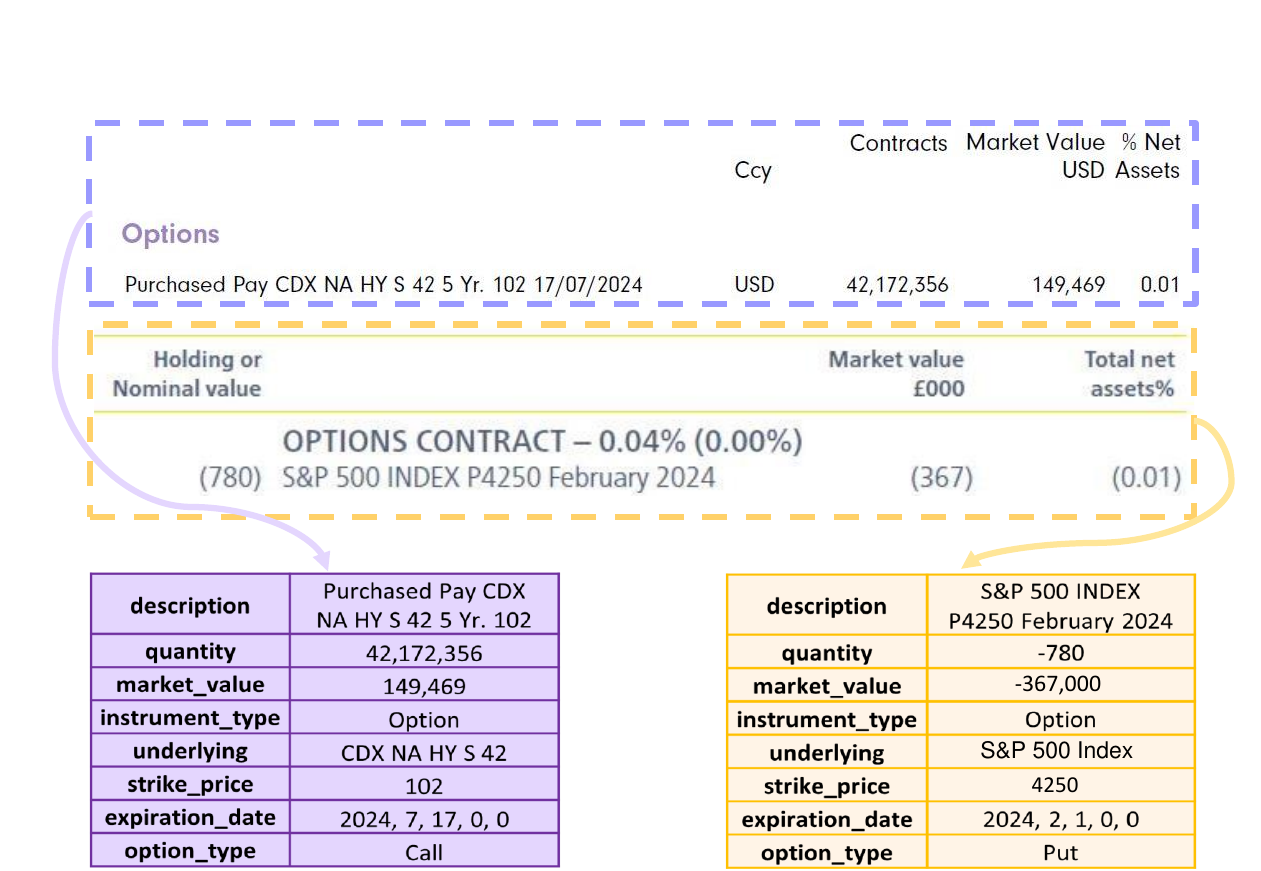}
\caption{\textbf{Complexity of Holdings Table in Regulatory Filings.} In the original format, multiple data attributes are displayed in a single line, with no bounding boxes, rendering the generation of structured outputs highly challenging. TASER enables the generation of structured outputs from highly variable, multi-page financial tables for complex instrument holdings. Negative quantities or market values denote short positions. See Appendix~\ref{app:exampleholdings} for additional outputs.}
\label{fig:norm_output}
\vspace{-4mm}
\end{figure}

\begin{figure}[th]
\centering
\includegraphics[width=0.97\linewidth, trim=0.65cm 0.7cm 0.97cm 0.7cm, clip]{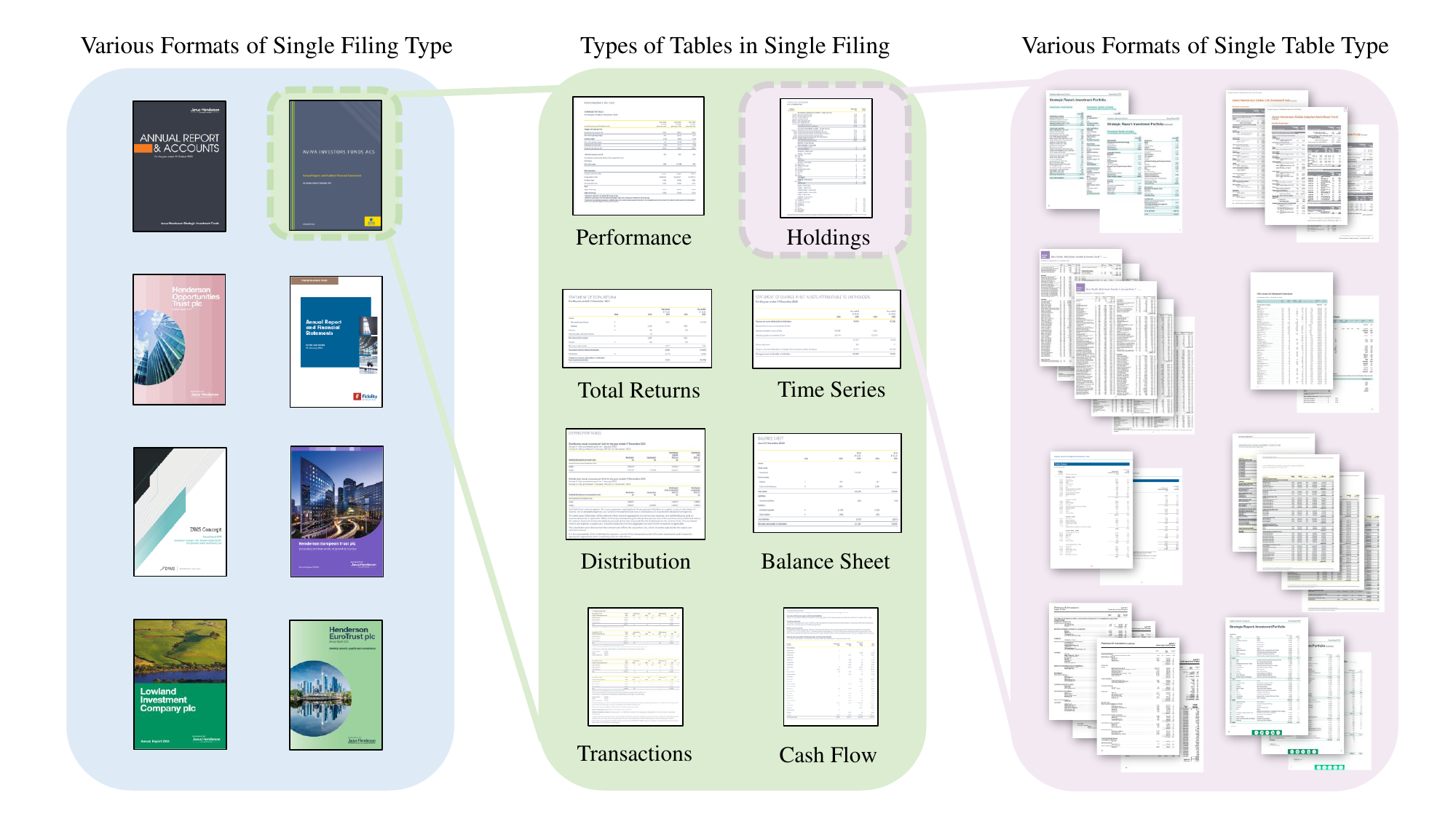}
\caption{\textbf{Variety and complexity of financial tables.} From leftmost column - for a single financial filing type, such as annual reports, there is no consistency among reports. Within each report, there are numerous table types with each type housing very different types of information. 
Even within a single table type (such as the Financial Holdings Table), there are numerous layout structures, as seen in the rightmost column. Due to the extreme heterogeneity of formatting, document layout, and table structure, traditional table extraction methods fail to perform for financial filings. }
\label{fig:doc-table complexity}
\vspace{-3mm}
\end{figure}

Financial documents, particularly annual regulatory filings for funds, house tables that govern \$68.9 trillion of investments globally \citep{ICI:2024}. By comparison, \$68.9 trillion is more than twice the total Gross Domestic Product (GDP) of the United States (\$29.1 trillion) \citep{WorldBankGDP}. This critical data is housed in the Financial Holdings Table (Figure \ref{fig:norm_output}), which outlines the entirety of an entity's investment holdings \citep{sec1934, EuropeanCommission}; this table has the highest row count (maximum 426 rows)--more than double the average row count of all other table types (Table \ref{tab:Financial Holdings Table Metrics}). These Financial Holdings Tables are long and highly heterogeneous in layout (Figure  \ref{fig:doc-table complexity}).
While generating structured outputs from these tables is critical for many regulatory and financial institutions to undertake basic QA (Question-Answering) tasks using an LLM (Large Language Model) or libraries such as \texttt{pandas} 
\citep{cho_fishnets}, there is a relative dearth of studies that focus on continual learning to extract from Financial Holdings Tables, compared to web or SQL tables \citep{Herzig_2020, pasupat-liang-2015-compositional, zhong2017seq2sqlgeneratingstructuredqueries}. 
Therefore, the following challenges exist in terms of parsing Financial Holdings Tables into structured, machine-readable outputs: \textbf{(1) One-to-many relationships between a document and the tables it houses} exacerbate standard model performance for table detection or structure recognition tasks. \textbf{(2) Financial Holdings Tables span across multiple pages}, rendering models that operate at the page level inefficient. \textbf{(3) Financial instruments are highly complex} with nested hierarchies. Therefore, details are often clumped in a single cell as seen in Figure \ref{fig:norm_output}.
\textbf{(4) Tabular layouts are heterogeneous with no bounding boxes}, mixing tables, text blocks, footnotes, and images, often without consistent labeling or alignment. 99.4\% of tables in our dataset, \textbf{TASERTab}, lack bounding boxes to efficiently identify a single cell. These challenges motivate our agentic table extraction methodology capable of goal-driven parsing and self-refinement, continuously learning and reasoning from errors.

\vspace{0.5mm}
\noindent\textbf{Contribution 1:} We propose a continuously learning, agentic table extraction methodology, \textbf{TASER} (Table Agents for Schema-guided Extraction and Recommendation) that performs detection, classification, extraction, and recommendation in a single pipeline by leveraging the schema invoked as a tool call. TASER is layout-agnostic and can operate for tables of any format. We compare our methodology against predominant methodologies and report TASER's 10.1\% improvement over Table Transformer \citep{yang2022tableformerrobusttransformermodeling} for detection.

\vspace{0.5mm}
\noindent\textbf{Contribution 2:} We demonstrate the effectiveness of our Recommender Agent to continuously improve the initial schema - reflecting a tunable and continuous self-learning loop. Throughout our training, we found that small batches are optimal for providing diverse and comprehensive recommendations to the original schema--however, at the cost of redundant recommendations. In contrast, large batches drive high precision recommendations at the cost of diversity. Thus, our results establish that self-learning via agents for table extraction is tunable; through adjusting batch size, we can control schema refinement to maximize actionable coverage while minimizing redundancy.

\vspace{0.5mm}
\noindent\textbf{Contribution 3:} We have constructed a manually labeled dataset \textbf{TASERTab} of ground truth labels for 3,213 real-world Financial Holdings Tables amounting to \$731.7B in value. We sourced the filings from fund websites, labeled the total net assets for each fund, and recorded the span of each Financial Holdings Table. We believe that this is the first dataset of its kind to provide access to real-world financial tables side by side with structured outputs.

\section{Related Work}

\label{sec:related}

\noindent\textbf{Information \& Table Extraction:}
Early information extraction relied on statistical models (HMMs~\citep{10.1145/375663.375682}, CRFs~\citep{10.5555/645530.655813}, heuristics~\citep{Columbia}, and graph-based layouts~\citep{liu2019graphconvolutionmultimodalinformation,qian-etal-2019-graphie,Meuschke_2023}, but still struggle with complex, heterogeneous tables. 

\vspace{0.5mm}
\noindent
\textbf{Table Representation Learning:}
Transformer-based table understanding and QA include TaPaS~\citep{Herzig_2020}, TaBERT~\citep{yin-etal-2020-tabert}, TaPEX~\citep{liu2022tapextablepretraininglearning}, TURL~\citep{deng2020turltableunderstandingrepresentation}, TUTA~\citep{Wang_2021}, and TableFormer~\citep{yang2022tableformerrobusttransformermodeling}. These methods encode text, structure, and layout, but few are benchmarked on long, dense, multi-page financial reports.

\vspace{0.5mm}
\noindent
\textbf{LLMs for Structured Data:}
General LLMs have strong performance for schema-conformant extraction via fine-tuning \& prompting~\citep{brown2020languagemodelsfewshotlearners,liu2024instructor}, while multimodal approaches (LayoutLM~\citep{Xu_2020,xu-etal-2021-layoutlmv2,huang2022layoutlmv3pretrainingdocumentai}, DONUT~\citep{kim2022ocrfreedocumentunderstandingtransformer}, DocFormer~\citep{appalaraju2021docformerendtoendtransformerdocument}, UniTable~\citep{peng2024unitableunifiedframeworktable}, and Table Transformer~\citep{smock2021pubtables1mcomprehensivetableextraction,carion2020endtoendobjectdetectiontransformers} improve layout awareness but still lag on long, fragmented tables~\citep{zhao2024tabpediacomprehensivevisualtable}.

\vspace{0.5mm}
\noindent\textbf{Financial Document Parsing:}
\citep{Watson_2020} has focused on table extraction from images while \citep{cho_fishnets} has focused on expert agent pipelines. Large-scale benchmarks such as DocILE~\citep{simsa2023docile}, BuDDIE~\citep{wang-etal-2025-buddie} have also focused on financial documents. 
\nocite{chi2019complicated}\nocite{smock2021pubtables1mcomprehensivetableextraction}\nocite{zhong2019image}\nocite{zhao2024tabpediacomprehensivevisualtable}\nocite{bizgraphqa}

\vspace{0.5mm}
\noindent\textbf{Agentic and Recursive Extraction:}
Recent methods cast LLMs as agents capable of iterative extraction and self-correction~\citep{shen2023hugginggptsolvingaitasks,smolagents,Watson_2023,yuan2025reinforcellmreasoningmultiagent}. Prompt-based feedback, introspective refinement, and episodic memory frameworks~\citep{madaan2023self,shinn2023reflexionlanguageagentsverbal,yao2023reactsynergizingreasoningacting} drive improvements in reasoning for complex extraction.


\section{Methodology}

\label{sec:method}

\subsection{System Architecture}
\begin{figure}[t]
    \centering
    \includegraphics[width=\columnwidth]{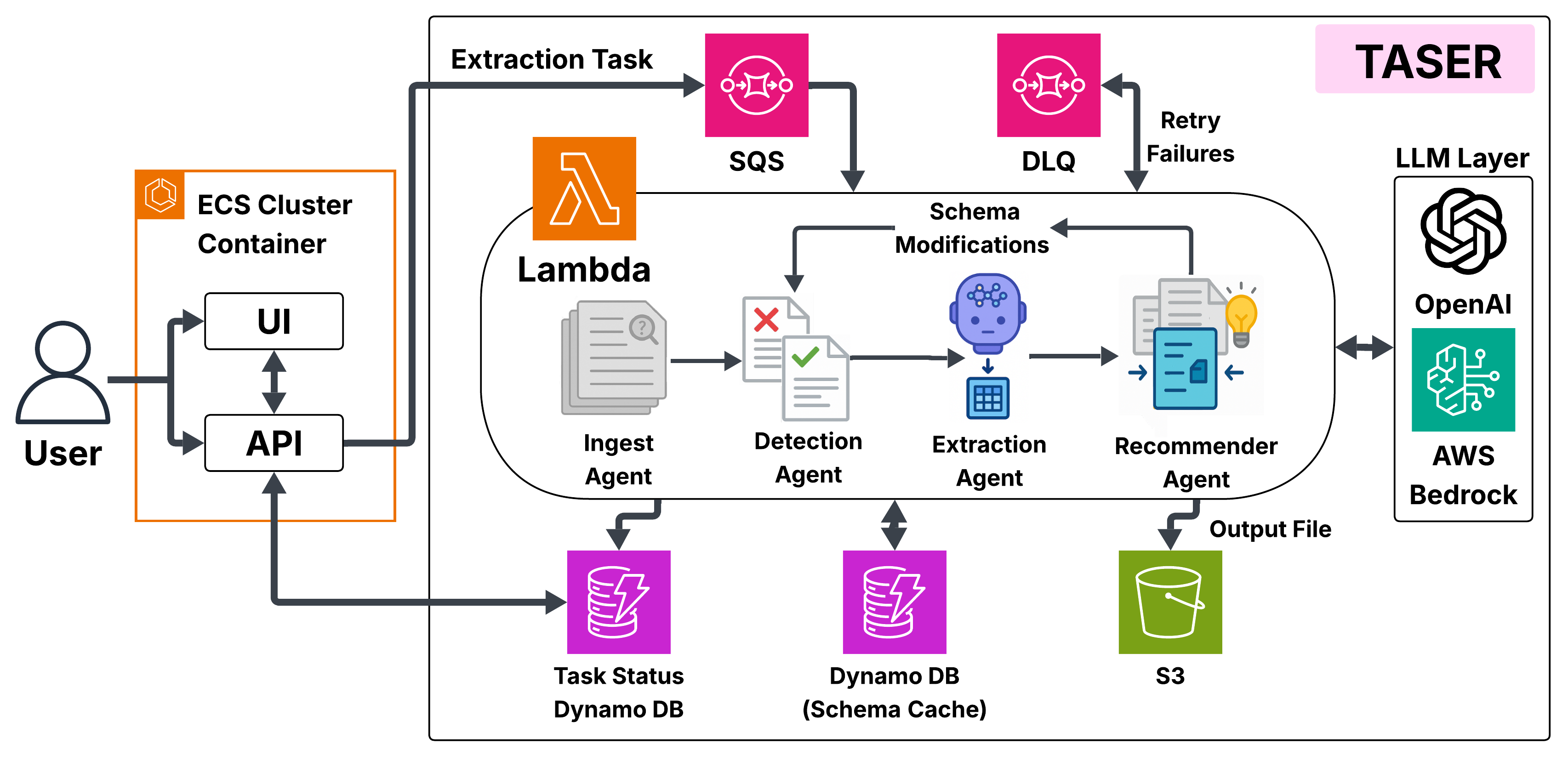} 
    \vspace{-5mm}
\caption{
    \textbf{TASER deployment architecture.} Users submit extraction requests through a UI or API hosted in an ECS container, which enqueues an extraction task to SQS. An AWS Lambda function orchestrates the Ingest, Detection, Extraction, and Recommender Agents, persisting task status and schema cache in DynamoDB and writing intermediary and final outputs to S3; failed tasks are routed to a dead-letter queue (DLQ) for later inspection. The Recommender Agent analyzes the intermediary output file and associated error stack to propose schema enhancement recommendations. Users can accept or reject these recommendations; accepted updates are written back to the schema cache and automatically retrigger the extraction pipeline, enabling TASER to continuously refine its extraction artifacts over time.
}
    \label{fig:llm-pipeline}
    \vspace{-3mm}
\end{figure}

Our methodology is composed of three core Large Language Model (LLM) agents, each with a distinct role. We conduct rigorous ablations to evaluate the importance of each agent. 
\begin{enumerate}[noitemsep, leftmargin=*, topsep=0pt, partopsep=0pt]
\item \textbf{Detector Agent:} Identifies candidate pages containing Financial Holdings Tables leveraging the initial schema provided. The prompt is tuned to maximize recall to avoid missing any Financial Holdings Tables. We provide our prompts in the Appendix (Figure \ref{figure:detection prompt}).
\item \textbf{Extractor Agent:}
Processes detected pages by prompting the LLM with the current \texttt{Portfolio} schema embedded in the prompt context. The LLM’s output is validated inline against the schema using Pydantic \& Instructor, producing a set of structured, type-checked instrument entries (Figure~\ref{fig:norm_output}).
\item \textbf{Recommender Agent:} Reviews unmatched extractions containing both false and true positives. A \emph{false positive} is a spurious extraction (e.g., headers/subtotals/footnotes/OCR noise or cells from non-holdings tables) that fails schema/type/consistency checks; a \emph{true positive} is a valid holdings field from a genuine row that the current schema cannot yet classify but passes those checks. The agent first filters false positives by re-validating each candidate under the current schema; it proposes schema modifications for the remaining true positives and, per class, recommends the minimal change needed.
\end{enumerate}
All agents interact through explicit artifacts: structured outputs, episodic error stacks, and schema definitions. Output validation is integrated into each agent's forward pass via Instructor~\citep{liu2024instructor}. TASER implements a recursive feedback loop, where errors and unmatched holdings identified in the initial extraction are escalated to the Recommender Agent, which provides recommendations to refine the schema and triggers re-extraction. This loop repeats until all entries are matched. 
A schematic of the full agentic pipeline is shown in Figure~\ref{fig:llm-pipeline}.

\subsection{Initial Schema Definition and Application}
TASER’s extraction process is anchored by an explicit, user-modifiable \texttt{Portfolio} schema that defines the target structure for Holdings Tables.
We implement this schema using Pydantic models; our initial schema reflects is informed by leveraging external knowledge \citep{sec1934}. Each schema consists of a base \texttt{Instrument} model, subclassed for common asset types (e.g., \texttt{Equity}, \texttt{Bond}, \texttt{Option}, \texttt{Swap}, \texttt{Forward}, \texttt{Future}, \texttt{Debt}, and an \texttt{Other} class for uncategorized rows. Each subclass specifies instrument-specific fields and validation logic (see App.~\ref{app:schema}).

\vspace{0.5mm}
\noindent
\textbf{Schema-Guided Extraction:} For each candidate page, the Extractor Agent prompts the LLM with the current schema embedded in the prompt context. The LLM is instructed to return a structured output, which is immediately parsed and validated against the schema using Pydantic’s type checking and validation logic. Outputs that fail schema validation (e.g., missing fields, type errors, or undeclared instruments) are flagged.

\vspace{0.5mm}
\noindent
\textbf{Schema Recommendations for Iterative Refinement:}
We formalize schema refinement as an iterative, LLM-driven clustering process that updates the schema to accommodate unmatched or novel holdings discovered during extraction \citep{novikov2025alphaevolvecodingagentscientific,zhang2023clusterllm}. At each iteration, the agent operates on the episodic error stack to propose schema modifications, and extraction is retried using the updated schema. This process continues until all entries are matched or no further improvements are possible.
Let $H = \{h_1, h_2, \ldots, h_N\}$ denote the set of unmatched holdings, and let $\Sigma^{(0)}$ be the initial schema. For each iteration $\ell$:
\begin{itemize}[noitemsep, leftmargin=*, topsep=0pt, partopsep=0pt, label={\tiny\raisebox{0.5ex}{$\blacktriangleright$}}]
\item $H^{(\ell)}$: Unmatched holdings at iteration $\ell$.
\item $\Sigma^{(\ell)}$: Current schema.
\item $g_\theta$: LLM-based schema suggestion function.
\item $B$: Batch size for error grouping.
\end{itemize}
The refinement loop (Algo.~\ref{alg:llm_schema_refinement}) proceeds as follows:
\begin{enumerate}[noitemsep, leftmargin=*, topsep=0pt, partopsep=0pt]
\item Partition $H^{(\ell)}$ into batches of size at most $B$.
\item For each batch, invoke $g_\theta$ with batch errors and $\Sigma^{(\ell)}$ to propose schema modifications.
\item Aggregate, cluster, and select recommendations
\item Update schema to $\Sigma^{(\ell+1)}$ and re-extract.
\item Update error stack and repeat until $H^{(\ell+1)}$ is empty or no new schema changes are suggested.
\end{enumerate}

\begin{figure}[t]
    \centering
    \includegraphics[width=0.9\columnwidth]{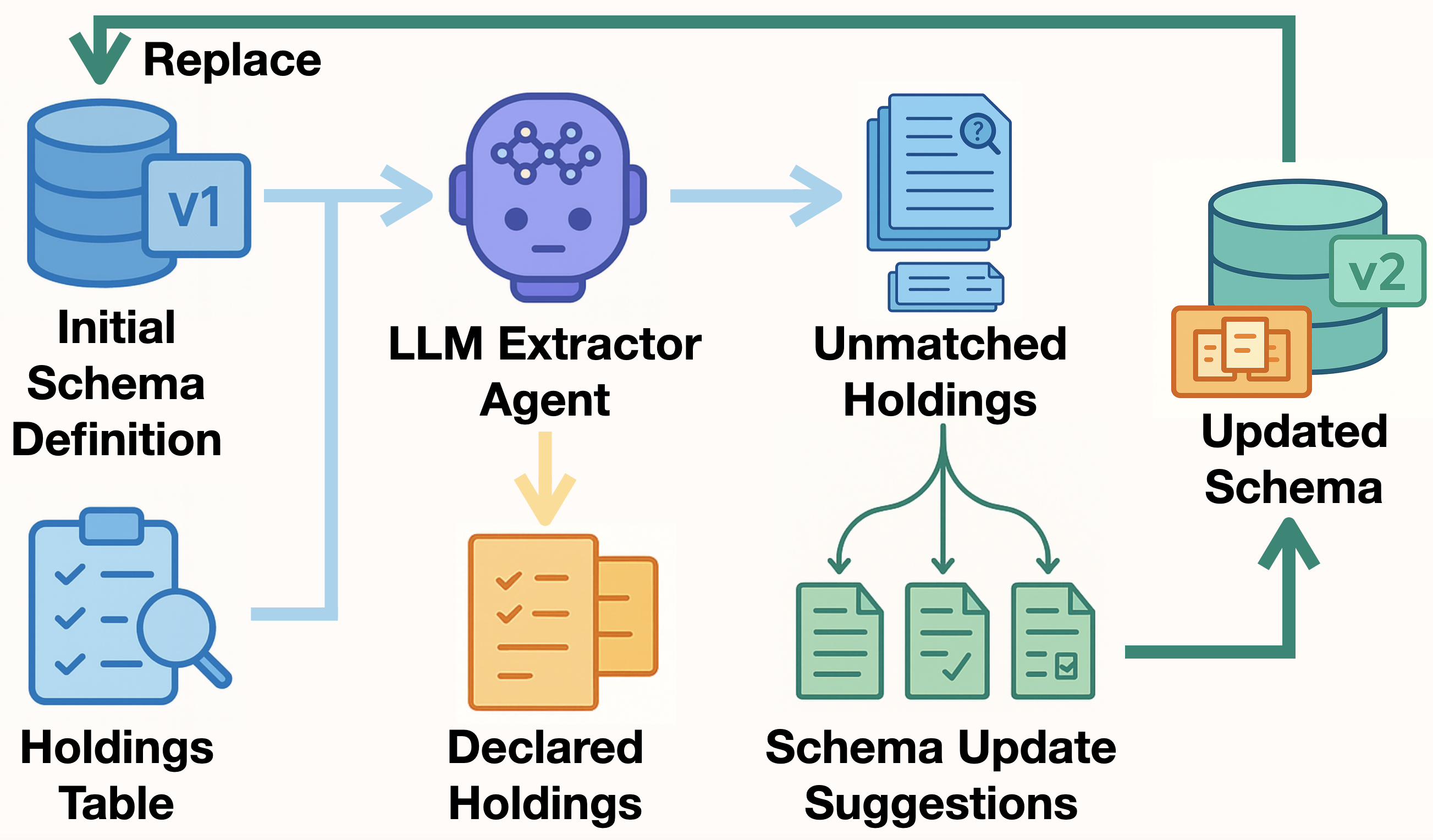}
    \vspace{-2mm}
    \caption{
        \textbf{Schema-Guided Agentic Refinement Loop.}
        The extraction pipeline begins with an \textit{Initial Schema Definition} (\texttt{v1}), which guides the LLM Extractor Agent as it processes the raw Holdings Table to produce \textit{Declared Holdings}. Holdings that do not match the schema are routed as \textit{Unmatched Holdings}, triggering the generation of \textit{Schema Update Suggestions}. These suggestions are reviewed, clustered, and aggregated by our Recommender Agent before updating the schema (\texttt{v2}), replacing the prior definition and closing the agentic feedback loop. This process enables continuous schema refinement and robust extraction.
    }
    \label{fig:schema-refinement-flow}
    \vspace{-3mm}
\end{figure}

\begin{algorithm}[t]
\small
\caption{LLM Iterative Schema Refinement}
\label{alg:llm_schema_refinement}
\begin{algorithmic}[1]
\REQUIRE Unmatched holdings $H = \{h_1, h_2, \ldots, h_N\}$, initial schema $\Sigma^{(0)}$, LLM schema suggestion function $g_\theta$, batch size $B$, stopping criterion $T$
\STATE Initialize $\ell \gets 0$
\STATE $H^{(0)} \gets H$ \COMMENT{Current unmatched holdings}
\STATE $\Sigma^{(0)} \gets$ initial schema
\WHILE{not stopping criterion $T$ met}
    \STATE Partition $H^{(\ell)}$ into batches $H^{(\ell)}_j$ of size at most $B$
    \STATE $S^{(\ell)} \gets \emptyset$ \COMMENT{Suggested schema modifications}
    \FOR{each batch $H^{(\ell)}_j$}
        \STATE $S_j^{(\ell)} \gets g_\theta(H^{(\ell)}_j, \Sigma^{(\ell)})$
        \STATE $S^{(\ell)} \gets S^{(\ell)} \cup S_j^{(\ell)}$
    \ENDFOR
    \STATE $S_{\text{selected}}^{(\ell)} \gets \text{AggregateAndSelect}(S^{(\ell)})$ \COMMENT{Aggregate suggestions}
    \STATE $\Sigma^{(\ell+1)} \gets \text{UpdateSchema}(\Sigma^{(\ell)}, S_{\text{selected}}^{(\ell)})$
    \STATE $H^{(\ell+1)} \gets$ \text{UnmatchedHoldings}$(H, \Sigma^{(\ell+1)})$
    \IF{$H^{(\ell+1)} = \emptyset$}
        \STATE \textbf{break}
    \ENDIF
    \STATE $\ell \gets \ell + 1$
\ENDWHILE
\STATE \textbf{return} $\Sigma^{(\ell+1)}$
\end{algorithmic}
\end{algorithm}

\subsection{Ablation Strategies and Efficiency}

We systematically ablate TASER to isolate the impact of schema-guided extraction, prompt engineering, and agentic feedback across four strategies:
\begin{enumerate}[noitemsep, leftmargin=*, topsep=0pt, partopsep=0pt]
\item \textbf{Raw Text Prompting:} The LLM is prompted only with the page text; extraction is based solely on a yes/no detection.
\item \textbf{Structured Chain-of-Thought (CoT):} Prompts include a minimal schema and few-shot examples, eliciting explicit reasoning traces before a final boolean decision.
\item \textbf{Full Schema Prompting:} The full \texttt{Portfolio} schema is embedded in the prompt, instructing the LLM to return structured, schema-conformant entries.
\item \textbf{Direct Schema Application:} The schema is directly applied to parsed page content without prior detection; extraction succeeds if any schema sub-model instantiates.
\end{enumerate}

\noindent
Table~\ref{tab:detection-metrics} reports detection and extraction via absolute dollar difference, and Table~\ref{tab:efficiency} compares computational efficiency in tokens and latency.

\begin{table*}[ht]
    \small
    \centering
    \resizebox{0.9\linewidth}{!}{
    \begin{tabular}{llcccc}
        \toprule
        \textbf{Provider} & \textbf{Model} & \textbf{Recall (\%)} & \textbf{Precision (\%)} & \textbf{F1 Score (\%)} & \textbf{Accuracy (\%)} \\
        \midrule
        Camelot    & Hybrid                   & 56.92 & 23.46 & 33.23 & 47.35 \\
        Microsoft & Table Transformer        & 99.76 & 32.75 & 49.31 & 46.27 \\
        \midrule
        OpenAI    & \texttt{gpt-4o-2024-11-20}      & \textbf{100.00} & 43.43 & 59.44 & 66.35 \\
        OpenAI    & \texttt{gpt-5-mini-2025-08-07}  & 94.92 & 54.15 & 68.96 & 80.33 \\
        OpenAI    & \texttt{gpt-4.1-2025-04-14}     & 95.80 & 54.32 & 69.33 & 80.49 \\
        OpenAI    & \texttt{gpt-5-nano-2025-08-07}  & 95.97 & 55.63 & 70.44 & 81.46 \\
        OpenAI    & \texttt{gpt-5-2025-08-07}       & 95.97 & 68.16 & \textbf{79.71} & \textbf{88.75} \\
        Anthropic & \texttt{claude\_sonnet-3-7}     & 88.97 & 57.34 & 69.73 & 82.22 \\
        Amazon    & \texttt{nova\_pro-v1-0}         & 85.90 & \textbf{69.45} & 76.84 & 88.07 \\
        \bottomrule
    \end{tabular}}
    \caption{\textbf{Detector performance across models on TASERTab.} Recall, precision, F1, and accuracy are reported for baselines and the Detector Agent instantiated with different LLMs. \texttt{gpt-4o-2024-11-20} attains perfect recall, while \texttt{gpt-5-2025-08-07} achieves the best overall F1 and accuracy. \texttt{nova\_pro-v1-0} delivers the highest precision but at the cost of lower recall, illustrating the trade-off between missing holdings tables and avoiding false positives.
    }
    \label{tab:model_performance}
    \vspace{-1.5mm}
\end{table*}

\begin{table*}[t]
\centering
\small
\resizebox{\linewidth}{!}{
\begin{tabular}{clcccccc}
\toprule
{} & \textbf{Method} & \textbf{Recall (\%)} & \textbf{Precision (\%)} & \textbf{F1 (\%)} & \textbf{Accuracy (\%)} & \textbf{TAD (USD)} & \textbf{Unaccounted} \\
\midrule
\multirow{4}{*}{\rotatebox{90}{\textbf{TASER}}} & 
(a) Raw Text Prompting           & \textbf{100.00} & 38.62 & 55.73 & 58.38 & \$ 107{,}066{,}845 & 0.015\% \\
& (b) Structured CoT               & \textbf{100.00} & 34.42 & 51.21 & 50.10 & \$ 120{,}577{,}458 & 0.016\% \\
& \textbf{(c) Full Schema Prompting}        & \textbf{100.00} & \textbf{43.43} & \textbf{59.44} & \textbf{66.35} & \underline{\$ 102{,}836{,}797} & \underline{0.014\%} \\
& (d) Direct Schema Application    & \textbf{100.00} & 41.84 & 58.30 & 63.99 & \$ 118{,}881{,}312 & 0.016\% \\
\bottomrule
\end{tabular}
}
\caption{\textbf{Detection and Extraction Performance Across Strategies.} While all TASER ablations achieve perfect recall, Full Schema Prompting yields the highest precision (43.43\%), F1 score (59.44\%), and overall accuracy (66.35\%), as well as the lowest total absolute difference (TAD) and unaccounted fraction, underscoring the value of embedding the complete Portfolio schema in the detection prompt.  Percentage of unaccounted holdings is out of \$731.7 billion (ground truth). Lower TAD and unaccounted percentages indicate higher dollar‐value fidelity.}
\label{tab:detection-metrics}
\vspace{-3mm}
\end{table*}

\begin{figure*}[ht]
    \centering
    \includegraphics[width=0.4\textwidth]{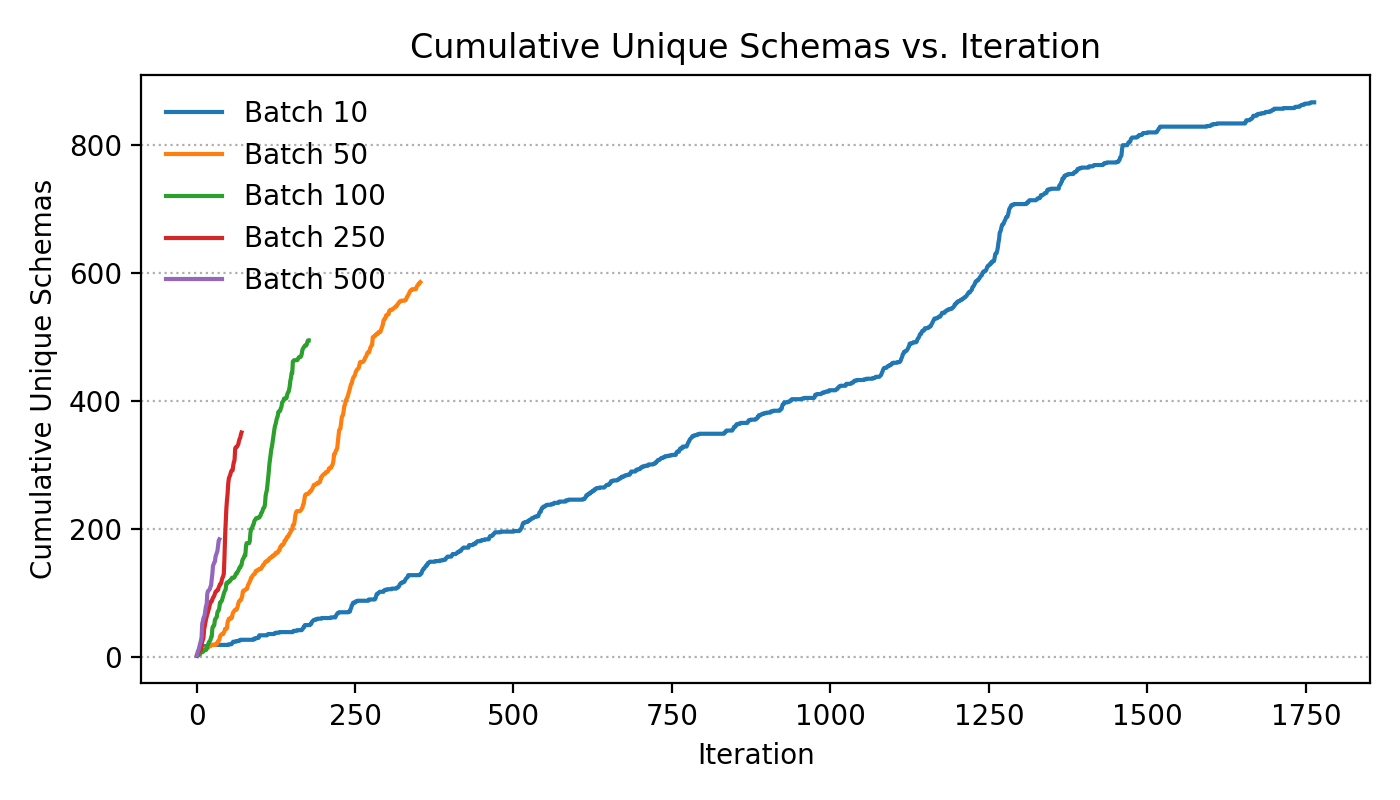}
    \includegraphics[width=0.4\textwidth]{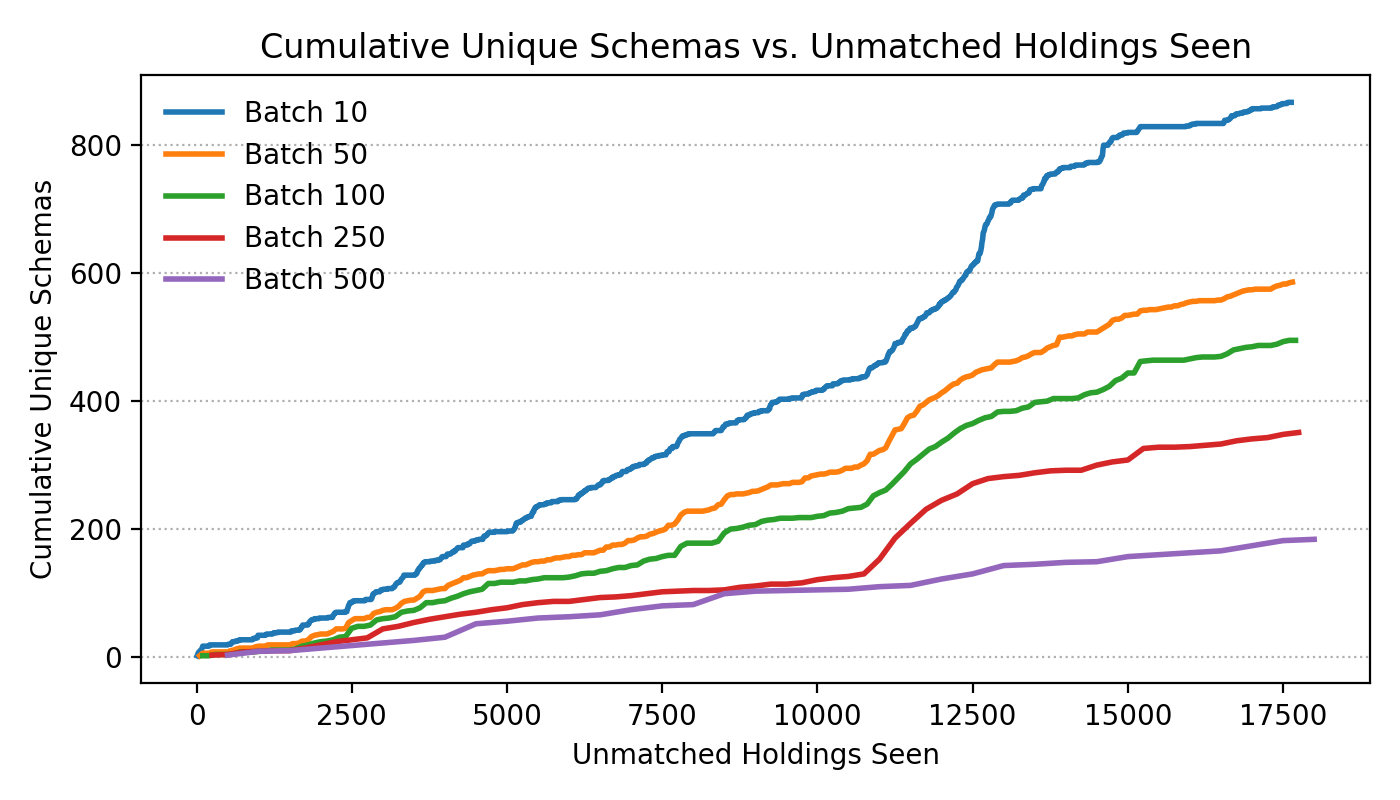}
    \vspace{-1mm}
    \caption{
    \textbf{Left:} Cumulative unique schemas per iteration; larger batches discover schemas rapidly but plateau quickly.
    \textbf{Right:} Cumulative unique schemas per unmatched holding seen; smaller batches ultimately yield more unique schemas but require more suggestions and generate more redundancy.
}
    \label{fig:unique-schemas-main}
    \vspace{-3mm}
\end{figure*}


\section{Experimental Setup}

\noindent\textbf{Detection Metrics:} We report \emph{recall}, \emph{precision}, \emph{F1}, and \emph{accuracy} for table detection, prioritizing recall to avoid missing Financial Holdings Tables.

\vspace{0.5mm}
\noindent\textbf{Extraction Metrics:} We assess extraction completeness by comparing TASER's outputs to ground truth labels. We manually label a total net asset value for each Holdings Table. We then compare this ground truth with our extractions, dubbed the \emph{total absolute difference (TAD)}.

\vspace{0.5mm}
\noindent\textbf{Schema Refinement Metrics:} 
\emph{Coverage} is the fraction of unmatched holdings aligned with at least one schema suggestion, using RapidFuzz string similarity with a lenient ($\geq 70$) threshold. We also report the number of new matched holdings after re-extraction with the suggested schemas added to \texttt{Portfolio}.
\emph{Diversity} is the average pairwise Levenshtein distance between suggestion attributes (name and generated schema). 
\emph{Collision rate} denotes the proportion of duplicate suggestions. 

\vspace{0.5mm}
\noindent
\textbf{Dataset and Model:}
We curate a diverse corpus totaling \textbf{22,584 pages}, \textbf{28M tokens}, and \textbf{\$731.7B} in holdings. 
Among \textbf{3,213 tables}, \textbf{57.53\%} exhibit hierarchical structure (via spanning cells). All Holdings Tables (\textbf{100\%}) are hierarchical.
While \textbf{39\%} of portfolios are single-page, \textbf{60.2\%} span multiple pages. The average length is \textbf{3.24 pages} ($\sigma=3.41$, max = 19). This variability underscores the need for multi-page detection and consolidation. 
Unless explicitly stated otherwise, all experiments use \texttt{gpt-4o-2024-11-20} as the LLM.

\section{Results and Discussion}

\subsection{Quantitative Evaluation}

\noindent\textbf{Detection:} Table~\ref{tab:detection-metrics} shows that all TASER ablations achieve perfect recall ($\sim$100\%), but precision ranges from 32.8\% (Table Transformer) up to 43.4\% (Full Schema Prompting), driving F1 scores between 49.3\% and 59.4\%. Embedding the full Portfolio schema in the prompt boosts precision by over 10\% relative to the vision‑only baseline and yields the highest F1 (59.4\%) and accuracy (66.4\%), demonstrating that in‑context schema guidance is critical. 

\vspace{0.5mm}
\noindent\textbf{Extraction:} Table~\ref{tab:detection-metrics} confirms schema-anchored extraction improves dollar‑value fidelity. Full Schema Prompting attains the lowest absolute difference (\$102.8M) and smallest unaccounted share (0.014\%), outperforming Raw Text Prompting (\$107.1M, 0.015\%) and Structured CoT (\$120.6M, 0.016\%). Direct Schema Application (skipping detection) incurs a higher error (\$118.9M; 0.016\%) by parsing spurious non‑holding pages. 

\subsection{Success Highlights}

\noindent\textbf{Cross-Document Consistency:} TASER classifies and extracts Holdings Tables despite varying titles (e.g., "Portfolio of Investments", "Schedule of Holdings", or "Investment Portfolio") and diverse structural formats. Despite the immense complexity of inputs, TASER consistently extracts and transforms these tables, ensuring that the final output appears as if sourced from a uniform set.

\vspace{0.5mm}
\noindent\textbf{Contextual Understanding:} TASER excels in handling contextual nuances, such as interpreting negative values denoted by parentheses (e.g., (140)) in zero-shot settings. Such domain-specific attributes are important for financial tables.

\vspace{0.5mm}
\noindent\textbf{Extracting Intricate Semantics:} TASER demonstrates a strong semantic understanding of financial terminology, which empowers it to extract accurately. For instance, TASER adeptly parsed the table entry ``GBP 4,700,000 | UK Treasury 0\% 19/02/2024 | 4668 | 1.48,'' correctly identifying the holding as a bond and extracting its attributes: \texttt{quantity}, \texttt{market value},  \texttt{coupon rate}, \texttt{maturity date}, and \texttt{issuer}.

\begin{figure}[t]
    \centering
    \includegraphics[width=0.9\linewidth]{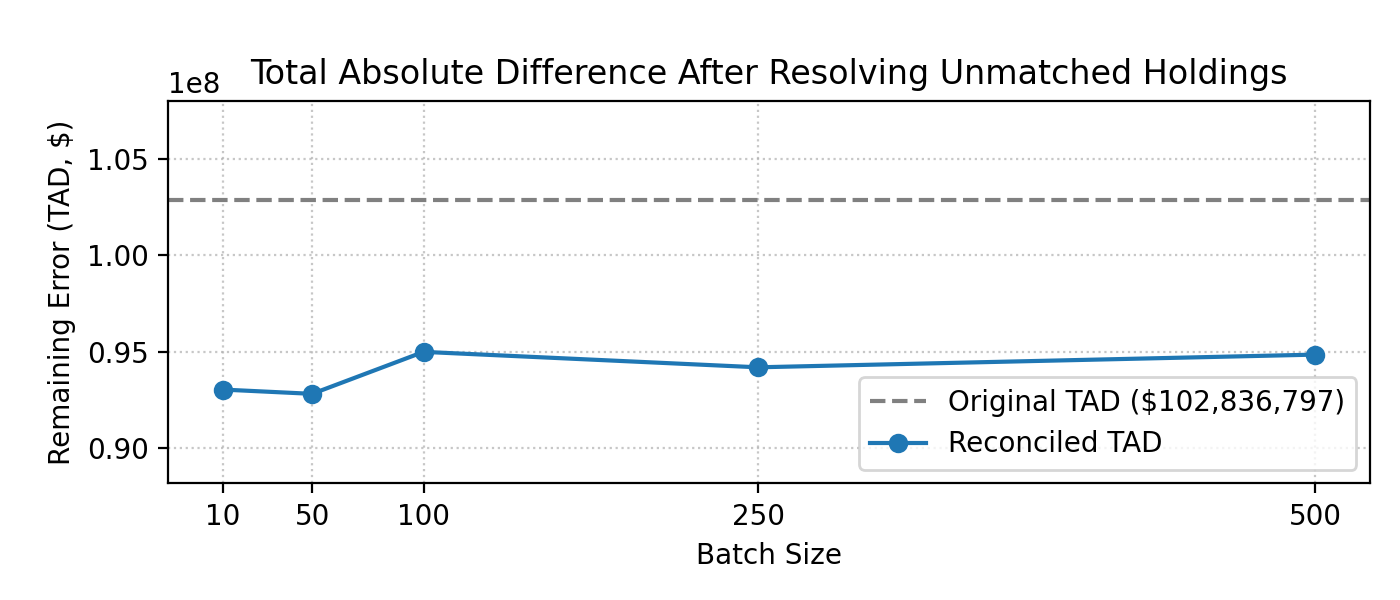}
    \vspace{-1mm}
    \caption{
        \textbf{Reduction in TAD after resolving unmatched holdings.}
        Remaining TAD is calculated after sequential reconciliation of unmatched holdings. 
        Error reduction is achieved by resolving the most significant unmatched holdings (Figs.~\ref{fig:lorenz}~\&~\ref{fig:heavy_tail}).
    }
    \label{fig:tad}
    \vspace{-3mm}
\end{figure}

\subsection{Batch Size Tradeoffs in Refinement}
Figure~\ref{fig:unique-schemas-main} (left) reveals that larger batch sizes (250, 500) rapidly expand the schema. However, this early acceleration comes at the cost of early saturation, after which few new unique schemas are discovered. In contrast, smaller batches require more iterations to reach the same number of unmatched holdings seen, but continue yielding unique schemas, resulting in the highest diversity when normalized by data processed (Figure~\ref{fig:unique-schemas-main}, right).
This improvement in coverage, however, is offset by increased redundancy. As shown in Appendix Figure~\ref{app:collisions}, smaller batches incur substantially more overlapping suggestions, reflecting a more granular and exploratory nature. Overall, these results highlight a key tradeoff: larger batches accelerate early discovery but plateau quickly, while smaller batches maximize cumulative schema diversity at the cost of redundancy and computation.
Our results establish that schema refinement via agentic feedback is both tractable and tunable. This indicates that an \textit{adaptive batching strategy} may be optimal: using larger batches to quickly identify high-yield schemas, followed by smaller batches for exhaustive diversity.

\vspace{0.5mm}
\noindent
\textbf{Schema Diversity and Utilization:}
Schema diversity, as measured by the average pairwise Levenshtein distance, is maximized for moderate batch sizes (100–250), as shown in Appendix~\ref{app:diversity}. 
While larger batch sizes (500) yield a higher proportion of utilized schemas—up to 59\%—smaller, more diverse batches tend to have lower utilization rates (Table~\ref{tab:schema-utilization}). Furthermore, the accretive gain in 402 additional unique schemas yielded only marginal improvements in holding coverage (6.1\%).
Figure~\ref{fig:coverage-utilization-tradeoff} illustrates this tradeoff: smaller batch sizes cover more unmatched holdings at the expense of efficiency (96.1\% coverage for 29.0\% utilization at batch size 10), whereas larger batches achieve higher schema utilization (59.0\% at batch size 500). 

\vspace{0.5mm}
\noindent\textbf{Improvements in TAD:}
Resolving the largest unmatched holdings yields a reduction in TAD of approximately 7–10\% across batch sizes, with the majority of improvement achieved by reconciling just the top 10–20\% of holdings (Table~\ref{tab:strict_tad_nav}). 

\vspace{0.5mm}
\subsection{Deployment of TASER} We outline the deployment architecture of TASER in Figure \ref{fig:llm-pipeline}, with additional system architecture details provided in
Appendix~\ref{app:taser-architecture}.

\vspace{-0.5mm}
\section{Conclusion}
\vspace{-1mm}
We present TASER for extracting complex Holdings Tables from documents through continual learning. Our high precision and recall across diverse layouts underscore the potential of agentic continual learning for financial table extraction. 

\section*{Disclaimer}
{
This paper was prepared for informational purposes by the Artificial Intelligence Research group of JPMorgan Chase \& Co. and its affiliates ("JPMorgan'') and is not a product of the Research Department of JPMorgan. JPMorgan makes no representation and warranty whatsoever and disclaims all liability, for the completeness, accuracy or reliability of the information contained herein. This document is not intended as investment research or investment advice, or a recommendation, offer or solicitation for the purchase or sale of any security, financial instrument, financial product or service, or to be used in any way for evaluating the merits of participating in any transaction, and shall not constitute a solicitation under any jurisdiction or to any person, if such solicitation under such jurisdiction or to such person would be unlawful.
}

\section*{Limitations}
Despite its strong performance, TASER remains susceptible to errors in low-resolution or scanned PDFs, where visual degradation can hinder accurate extraction. Ambiguities in financial documents, such as undefined asset classes or implicit references, pose challenges that cannot always be resolved without external knowledge or manual intervention. While recursive prompting enhances completeness, it introduces added latency and computational overhead. Additionally, TASER relies on prompt-based weak supervision due to the lack of fine-grained, labeled datasets for complex instrument types, which may limit generalization. Finally, TASER does not yet model interactions between table rows or instrument relationships beyond the schema level, which may affect downstream tasks such as portfolio risk analysis or exposure aggregation.

\bibliography{custom}

\clearpage

\appendix

\begin{table*}[t]
\centering
\small
\begin{tabular}{lccccc}
\toprule
\textbf{Model} & \textbf{Modality} & \textbf{Primary Task} & \textbf{Promptable}  \\
\midrule
Camelot  
  & Vision + Spatial  
  & Heuristic Table Detection \& Parsing  
  & No  
  \\
Table Transformer 
  & Vision  
  & Detection \& Structure Recognition 
  & No  
  \\
TaPas 
  & Text  
  & Table‑based QA  
  & Partially  
  \\
TAPEX 
  & Text  
  & Programmatic Extraction (SQL)  
  & Partially  
  \\
  \midrule
\textbf{TASER (ours)}  
  & Vision + Text  
  & Schema‑guided Extraction  
  & Yes  
  \\
\bottomrule
\end{tabular}
\caption{\textbf{Comparison of representative table extraction and reasoning models.} 
Our work extends prior methods by introducing a fully agentic, schema-guided extraction framework for highly complex financial tables, leveraging prompt-based self-refinement and continuous schema adaptation.
}
\label{tab:model-comparison}
\vspace{-1mm}
\end{table*}

\begin{table*}[t!]
\centering
\small
\begin{tabular}{clcccc}
\toprule
{} & \textbf{Method} & \textbf{Recall (\%)} & \textbf{Precision (\%)} & \textbf{F1 (\%)} & \textbf{Accuracy (\%)} \\
\midrule
\multirow{4}{*}{\rotatebox{90}{\textbf{Camelot}}}
& Stream              & 28.02 & 17.56 & 21.59 & 53.16 \\
& Lattice             & 14.01 & 12.72 & 13.33 & 58.08 \\
& Network             & 42.62 & 21.50 & 28.58 & 50.97 \\
& Hybrid              & 56.92 & 23.46 & 33.23 & 47.35 \\
\midrule
& Table Transformer \citep{smock2021pubtables1mcomprehensivetableextraction} & 99.76 & 32.75 & 49.31 & 46.27 \\
\midrule
\multirow{4}{*}{\rotatebox{90}{\textbf{TASER}}} 
& (a) Raw Text Prompting           & 100.0 & 38.62 & 55.73 & 58.38 \\
& (b) Structured CoT               & 100.0 & 34.42 & 51.21 & 50.10 \\
& (c) Full Schema Prompting        & 100.0 & \textbf{43.43} & \textbf{59.44} & \textbf{66.35} \\
& (d) Direct Schema Application    & 100.0 & 41.84 & 58.30 & 63.99 \\
\bottomrule
\end{tabular}
\caption{\textbf{Detection performance across all benchmarked strategies.} Camelot variants underperform across all metrics, with Hybrid achieving the highest F1 score (33.23\%) among them. TASER consistently achieves perfect recall and outperforms both Camelot and Table Transformer baselines, with Full Schema Prompting yielding the best precision (43.43\%), F1 score (59.44\%), and accuracy (66.35\%).}
\label{tab:detection-metrics-appendix}
\end{table*}

\section{TASER System Architecture}
\label{app:taser-architecture}

\subsection{Overview}
TASER is an event-driven microservice architecture composed of several semi-independent agents, such as text extraction and table detection modules. 
Each agent operates on a shared cloud infrastructure, which typically consists of an AWS Lambda function that listens to a queue (Amazon SQS), manages file operations in Amazon S3, and utilizes a dead-letter queue (DLQ) for retry handling.
 
\subsection{User Interaction and Deployment}
Users may interact with TASER either through a graphical user interface (UI) or directly via the application programming interface (API). 
Both the UI and API are deployed within an Amazon ECS cluster, with horizontal and vertical scaling enabled to respond to user demand. 
The UI is updated by periodically polling the API for \textit{task status}, which is
maintained in Amazon DynamoDB. 


\subsection{Task Cache}
An additional component is the \textit{task cache}, implemented using DynamoDB. 
The primary purpose of the cache is to reduce latency and operational costs by retrieving previously computed results. 
For all agents, the cached value is typically an S3 location indicating where the result of a prior execution is stored. 
The cache key varies by agent but generally includes model-specific information (e.g., model name, temperature settings) and the prompt used for generation.
A cache entry for the table detection agent is structured as follows:

\begin{center}
\small
\begin{tabular}{@{}ll@{}}
  \toprule
  \textbf{Key:}   & \texttt{\{schema\_id, prompt\_id, document\_text\_id,} \\
                  & \texttt{model\_name, model\_configs\}} \\
\midrule
  \textbf{Value:} & S3 location \\
  \bottomrule
\end{tabular}
\normalsize
\end{center}



\subsection{Design Rationale: API-Centric Orchestration}
A core design decision for TASER was to utilize an API-centric approach for orchestrating workflows. An alternative would have been to configure each agent to automatically trigger an AWS Lambda function upon the arrival of a new file in S3 (either by directly configuring the S3 bucket or using AWS EventBridge). While this method is feasible, it means that any change to workflows would require reprovisioning AWS infrastructure.
In contrast, with the current architecture, redesigning a workflow is
as simple as refactoring the UI or API client code (for example,
changing the order of API calls) and redeploying the application,
without modifying the underlying cloud infrastructure.

\section{Camelot Table Parsing Modes}
\label{sec:camelot}

For completeness, we also compare TASER’s detection performance against the four table detection modes in Camelot\footnote{\url{https://github.com/camelot-dev/camelot}}. The best-performing variant (Hybrid) achieves an F1 score of 0.33, still below TASER’s weakest ablation (0.51). Full results are presented in Table~\ref{tab:detection-metrics-appendix}.
Note that our financial tables primarily consist of unruled, whitespace-separated tables with alignment-based structure. Below is a brief summary of each mode:
\begin{itemize}[noitemsep, leftmargin=*, topsep=0pt, partopsep=0pt, label={\tiny\raisebox{0.5ex}{$\blacktriangleright$}}]
    \item \textbf{Stream:} Groups text using whitespace and y-axis alignment. Suitable for unruled tables, but yielded low precision on our data ($F1=21.6\%$).
    
    \item \textbf{Lattice:} Uses image-based line detection to extract ruled tables. Less effective for our dataset due to the rarity of bordered layouts ($F1=13.3\%$).
    
    \item \textbf{Network:} Detects tables via text alignment patterns using bounding boxes. Performs better on our format, which lacks explicit ruling ($F1=18.6\%$).
    
    \item \textbf{Hybrid:} Combines Network's structure with Lattice's grid refinement. Achieved the highest F1 score (33.23\%) among Camelot modes, confirming the benefit of integrating both visual and alignment cues.
\end{itemize}

\begin{figure}[t]
    \centering
    \includegraphics[width=0.48\textwidth]{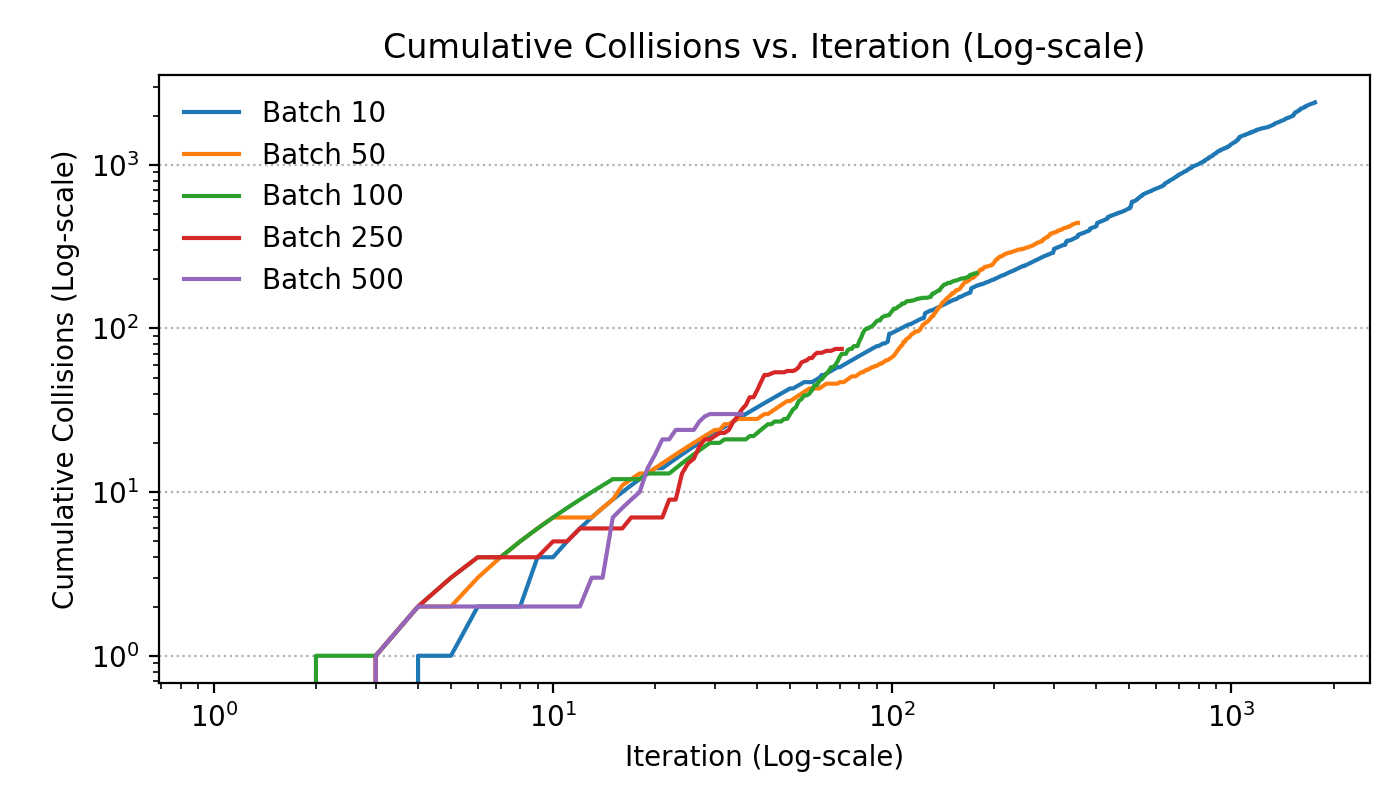}
    \includegraphics[width=0.48\textwidth]{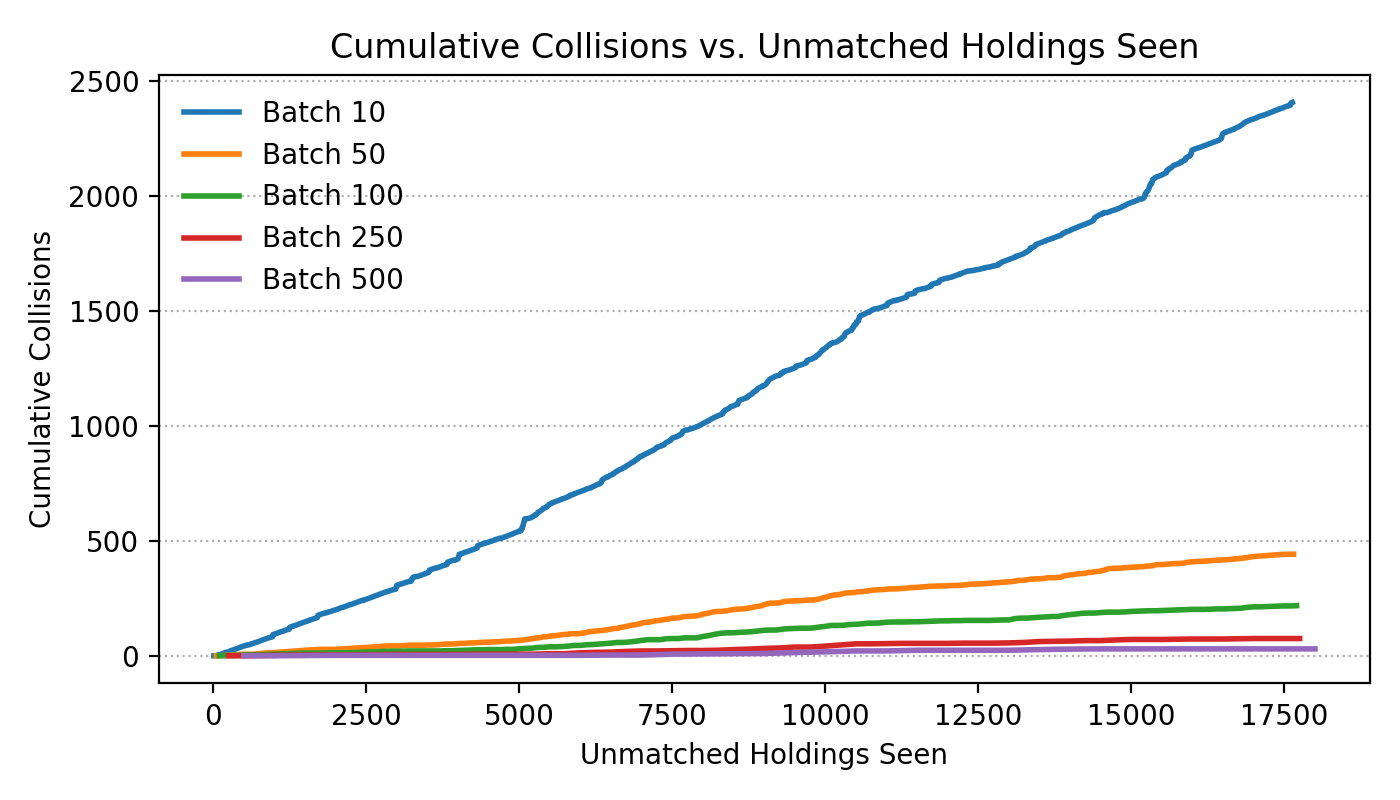}
    \caption{
    \textbf{Cumulative collisions per unmatched holding;} smaller batches incur more collisions, reflecting greater redundancy.
}

    \label{app:collisions}
\end{figure}

\section{TASER Annotation Process}
We manually sourced each financial document directly from the fund entity’s public website, ensuring broad coverage across instrument types. Annotations were performed at the page, table, and holdings level (which may span hundreds of pages). For every filing and fund, we recorded the page-span for the portfolio of investments table and the net asset value across all holdings for that fund.

\section{TASER Dataset Release}
TASER is built on public fund documents. Our release will include labels for the positions of holdings tables, the recorded net asset value, the fund name, multi-page spans, and a URL reference to the public fund document. Each pdf filing is hosted by the fund's advisor, as required by regulation.

\begin{table}[t]
\small
\centering
\begin{tabular}{r r r r}
\toprule
\textbf{Batch} & \textbf{Remaining} & \textbf{TAD Reduction} & \textbf{NAV} \\
\textbf{Size} & \textbf{TAD (\$)} & \textbf{(\%)} & \textbf{Extracted (\$)} \\
\midrule
500 & 94,843,638 & 7.8\% & 7,993,158 \\
250 & 94,185,693 & 8.4\% & 8,651,103 \\
100 & 95,985,588 & 6.7\% & 7,851,209 \\
50  & 92,781,421 & 9.8\% & 10,025,376 \\
10  & 93,032,549 & 9.6\% & 9,804,248 \\
\bottomrule
\end{tabular}
\caption{
\textbf{Remaining Total Absolute Difference} (TAD, \$) and Net Asset Value (NAV, \$) extracted from reconciled unmatched holdings by batch size.
}
\label{tab:strict_tad_nav}
\end{table}

\section{Document Preprocessing}

For each PDF filing, TASER extracts raw text, layout metadata, and embedded images using a hybrid pipeline based on \texttt{pdfplumber}. Each page is parsed into normalized text blocks and layout primitives, preserving spatial relationships and read order. Minimal normalization is applied, including Unicode cleanup and header/footer removal. Each page object includes:
\begin{itemize}[noitemsep, leftmargin=*, topsep=0pt, partopsep=0pt, label={\tiny\raisebox{0.5ex}{$\blacktriangleright$}}]
\item Raw text blocks (reading order preserved)
\item Bounding boxes and font metadata
\item Embedded images (if any)
\end{itemize}
We apply Unicode normalization (NFKC), whitespace collapse, and filter out repeated headers/footers via regex matching. Optionally, OCR is performed if text extraction fails. Code and parameters are available upon request.

\begin{figure}[t]
    \centering
    \includegraphics[width=0.99\linewidth]{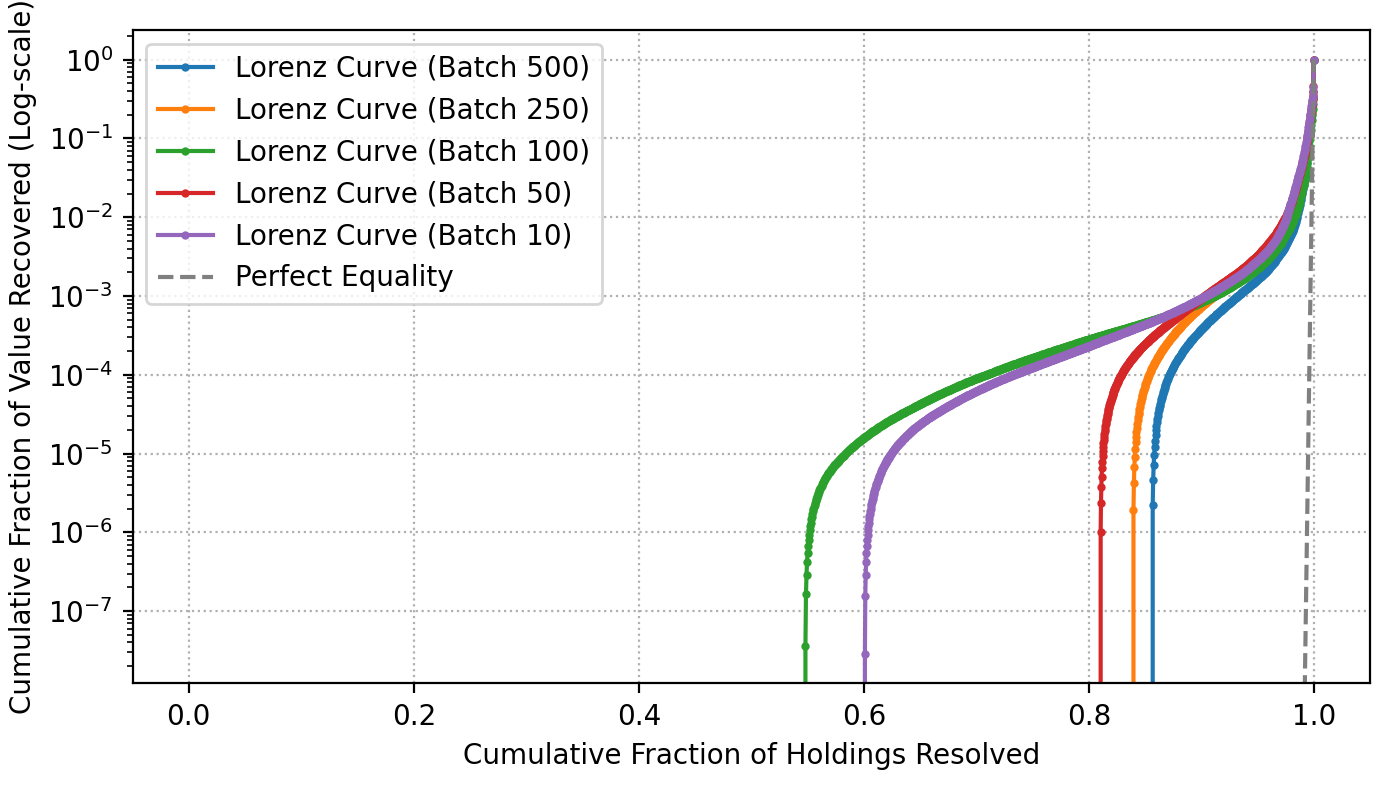}
    \caption{
        \textbf{Heavy-tailed distribution of value recovery from unmatched holdings across batch sizes.} 
        We report the Lorenz curves for the cumulative fraction of value recovered as a function of the fraction of “other” holdings resolved. 
        For all batch sizes, a small number of matches account for the vast majority of recovered net asset value, while most resolved holdings contribute negligibly. 
        The bow of each curve away from the diagonal illustrates the extreme concentration of recoverable value in the “head,” characteristic of a heavy-tailed regime.
    }
    \label{fig:lorenz}
\end{figure}

\begin{figure}[t]
    \centering
    \includegraphics[width=0.99\linewidth]{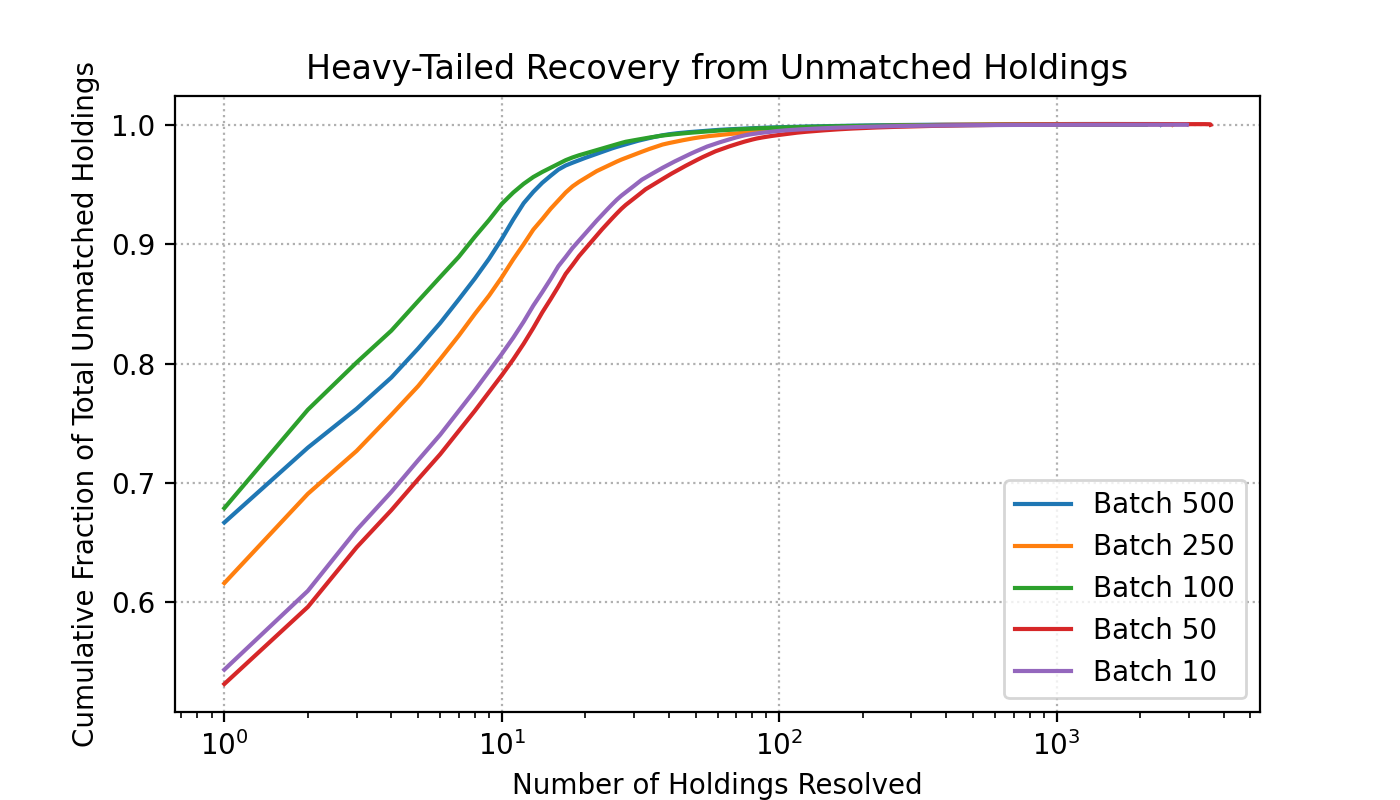}
    \caption{
        \textbf{Cumulative recovery fraction vs.\ number of holdings resolved.}
        Cumulative fraction of total value recovered as a function of the number of unmatched holdings resolved (log-log scale).
        The steep initial rise for each batch size indicates that the largest recoveries are concentrated among the first few resolved holdings; subsequently, improvement plateaus, indicating diminishing returns from resolving additional holdings in the long tail.
    }
    \label{fig:heavy_tail}
\end{figure}

\section{Parallelization and Fund Construction}

\noindent
\textbf{Extraction:}
To efficiently process large, multi-page filings, TASER employs parallelization (20 workers) at both the document and page levels. Each agent operates asynchronously across document batches: Detector and Extractor agents process candidate pages in parallel, while the Recommender agent operates downstream on the resulting artifacts.

\noindent
\textbf{Merging:}
For fund-level construction, extracted tables from consecutive pages are merged deterministically. Entity resolution is performed by matching predicted fund names and table headings across pages, while units and currencies are normalized to a consistent reporting standard through a boolean flag \texttt{value\_in\_thousands}. Partial extractions are reconciled using strict types in the response model, whose validation errors re-prompt the LLM on specific extraction errors to ensure a unified, schema-conformant portfolio representation for each fund.

\begin{table*}[t]
\small
\centering
\begin{tabular}{lrrrr}
\toprule
{} & \textbf{Avg. \# Tables}  & \textbf{Avg. Rows} & \textbf{Avg. Columns} & \textbf{Avg. Spanning} \\
 \textbf{Dataset}
 & \textbf{Per Topology} 
 & \textbf{Per Table}
 & \textbf{Per Table}
 & \textbf{Cells per Table}
\\
\midrule
SciTSR        & 5.70  & 9.28   & 5.19  & 0.77 \\
PubTabNet     & 4.13  & 14.05  & 5.39  & 2.24 \\
FinTabNet     & 11.80 & 11.93  & 4.36  & 1.01 \\
PubTables-1M  & 3.78  & 13.41  & 5.46  & 3.01 \\
\midrule
\textbf{TASERTab} & \textbf{11.00} & \textbf{53.70} & \textbf{6.36} & \textbf{2.67} \\
\bottomrule
\end{tabular}%
\caption{\textbf{Complexity of table instances across datasets.} TASERTab exhibits almost five times the number of rows compared to other datasets. The maximum row count in TASERTab is 426 rows across 44 pages for a single Financial Holdings Table.}
\label{tab:table-complexity}
\end{table*}

\begin{table*}[t]
\small
\centering
\resizebox{0.97\linewidth}{!}{%
\begin{tabular}{lrrrrrr}
\toprule
{} 
&  
& \textbf{\# Unique Cell}  
& \textbf{Avg. \# Tables}  
& \textbf{Avg. Rows} 
& \textbf{Avg. Columns} 
& \textbf{Maximum Page Span} \\
 \textbf{Dataset}
 & \textbf{\# Tables}
 & \textbf{Topologies} 
 & \textbf{Per Topology} 
 & \textbf{Per Table}
 & \textbf{Per Table}
 & \textbf{}
 \\
\midrule
Financial Holdings Table        & 1933    & 621     & 3.11  & \textbf{53.7}   & 6.36  & 44 \\
All Other Tables     & 1280   & 331  & 4.32 & \textbf{26.9}  & 3.87 & 37 \\
\bottomrule
\end{tabular}%
}
\caption{\textbf{Complexity of Financial Holdings Tables}}
\label{tab:Financial Holdings Table Metrics}
\end{table*}

\section{Schema Definitions and Portfolio Model}
\label{app:schema}
\begin{figure*}[t]
\centering
\includegraphics[width=0.93\textwidth,trim=1.9cm 7.4cm 5.5cm 2.05cm, clip]{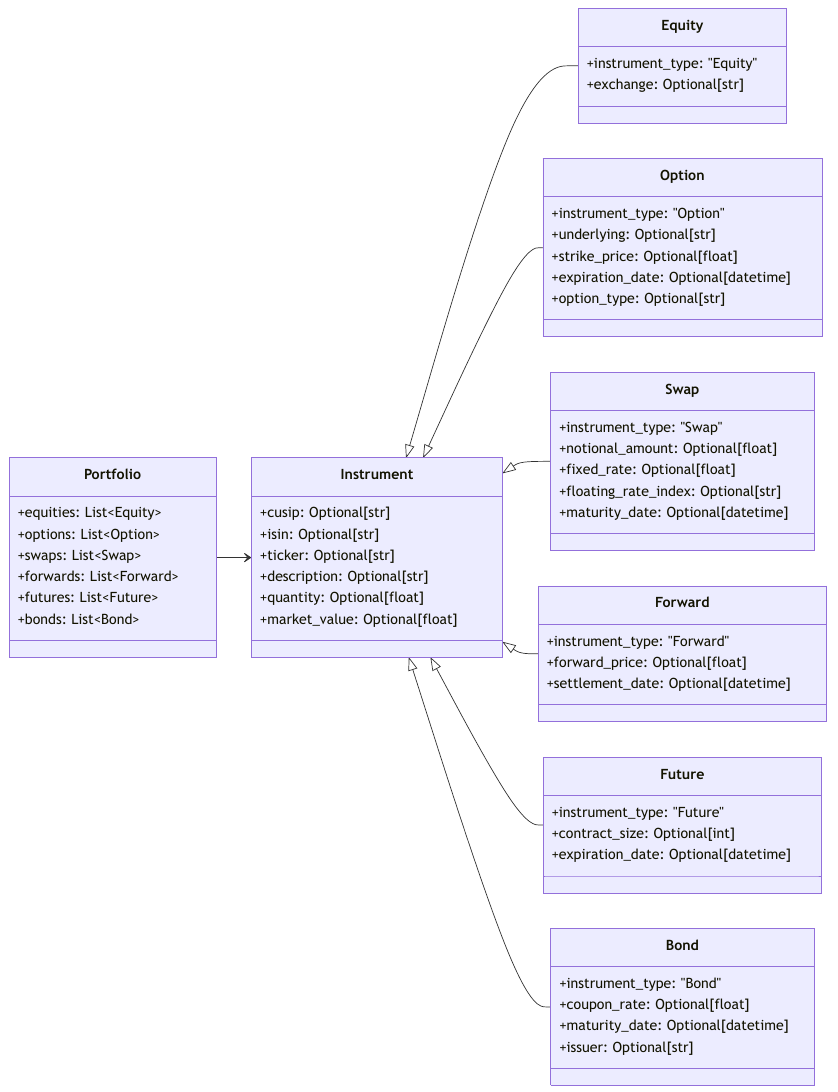}
\caption{\textbf{Class diagram of the initial \texttt{Portfolio} schema,} showing the top‑level \texttt{Portfolio} containing a collection of \texttt{Instrument} objects, each subclassed into specific security types (\texttt{Equity}, \texttt{Bond}, \texttt{Future}, \texttt{Forward}, \texttt{Swap}, \texttt{Option}) to capture their unique attributes.}
\label{fig:portfolio}
\vspace{-3mm}
\end{figure*}

\paragraph{Portfolio Base Model:}
The core \texttt{Instrument} base model in our Pydantic model is subclassed into the following classes (see Figure~\ref{fig:portfolio} for the full class diagram).  This is our initial schema composed of some of the most well-known financial instruments:
\begin{itemize}[noitemsep, leftmargin=*, topsep=0pt, partopsep=0pt, label={\tiny\raisebox{0.5ex}{$\blacktriangleright$}}]
  \item \textbf{\texttt{Equity}}: a share of ownership in a corporation, representing residual claims on earnings and assets.
  \item \textbf{\texttt{Bond}}: a fixed‐income security issued by governments or corporations, paying periodic coupons and returning principal at maturity.
  \item \textbf{\texttt{Future}}: an exchange‐traded contract obligating the buyer or seller to transact an asset at a predetermined price on a specified future date.
  \item \textbf{\texttt{Forward}}: an over‐the‐counter agreement to buy or sell an underlying asset at a set price on a future date, customizable but counterparty‐risky.
  \item \textbf{\texttt{Swap}}: a bilateral contract to exchange cash flows (e.g., fixed vs.\ floating interest rates or different currencies), with terms set at initiation.
  \item \textbf{\texttt{Option}}: a derivative granting the right, but not the obligation, to buy (call) or sell (put) an underlying asset at a specified strike price before or at expiry.
  \item \textbf{\texttt{Debt}}: a broad class of fixed-income securities including variable return notes, medium-term notes, and government bonds, not otherwise classified as standard bonds.
  \item \textbf{\texttt{Equity Linked Note (ELN)}}: a structured product whose returns are linked to the performance of an underlying equity or basket of equities.
  \item \textbf{\texttt{Other}}: a catch-all for instrument types not covered by the above classes, enabling schema extension and novelty detection.

\end{itemize}

\begin{figure}[t]
    \centering
    \includegraphics[width=0.98\columnwidth]{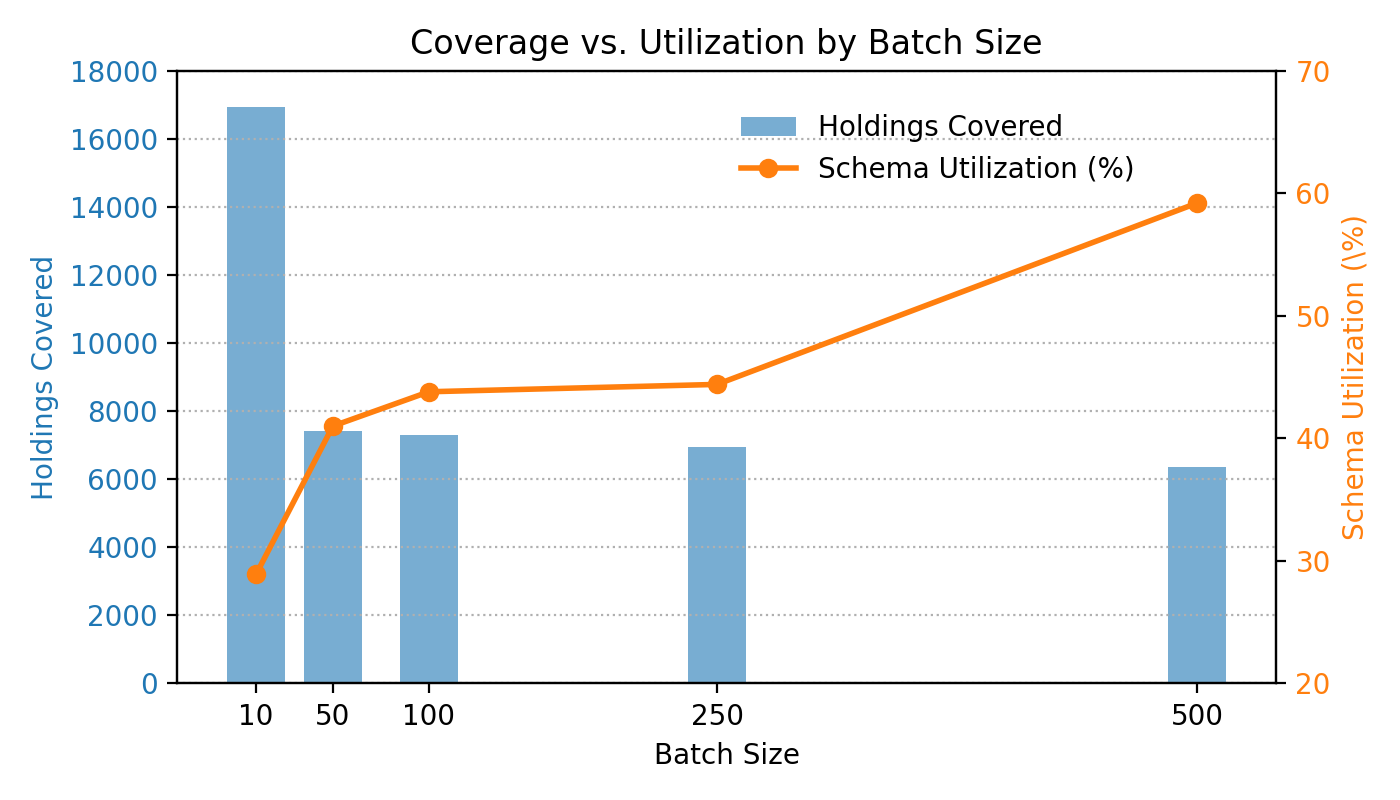}
    \caption{
        \textbf{Coverage vs. Utilization by Batch Size.}
        The number of unmatched holdings covered (bars, left axis) decreases with increasing batch size, while the fraction of schema suggestions utilized (line, right axis) increases. This highlights a tradeoff: small batches are more exhaustive in coverage, but large batches are more efficient—yielding fewer “wasted” schema suggestions.
    }
    \label{fig:coverage-utilization-tradeoff}
\end{figure}

\begin{table}[!ht]
\centering
\resizebox{\linewidth}{!}{
\begin{tabular}{c|ccc|ccc}
\toprule
\textbf{Batch} & \multicolumn{3}{c|}{\textbf{Name Diversity}} & \multicolumn{3}{c}{\textbf{Schema Diversity}} \\
\textbf{Size}  & \textbf{Avg} & \textbf{Min} & \textbf{Max} & \textbf{Avg} & \textbf{Min} & \textbf{Max} \\
\midrule
10  & 25.94 & 0 & 82 & 331.80 & 0 & 1387 \\
50  & 22.67 & 0 & 78   & 313.41 & 0 & 1305 \\
100 & 24.21 & 0 & 71   & 350.32 & 0 & 1569 \\
250 & 22.60 & 0 & 55   & 342.97 & 0 & 1230 \\
500 & 20.35 & 0 & 54   & 246.40 & 0 & 737 \\
\bottomrule
\end{tabular}
}
\caption{Diversity metrics of unique schema suggestions for varying batch sizes. We report the average/minimum/maximum pairwise Levenshtein distance; “schema” metrics are over the entire generated schema, “name” is on the generated holding class name.}
\label{tab:diversity}
\end{table}

\section{Ablation Strategies}

\paragraph{Raw Text Prompting.}
For the baseline ablation, we prompt the LLM solely with the raw page text, asking whether a portfolio table is present via a simple yes/no detection prompt. Upon affirmative detection, the LLM is instructed to extract a portfolio table from the same text, returning the result as a structured object with a \texttt{portfolio} field, but without access to any schema or structural guidance. This strategy measures the LLM's extraction performance in the absence of schema scaffolding or explicit reasoning.

\paragraph{Structured Chain-of-Thought (CoT).}
To assess the impact of explicit reasoning on table detection, we prompt the LLM with the page text and require a structured Pydantic output containing both a chain-of-thought explanation (\texttt{table\_chain\_of\_thought}) and a boolean indicating the presence of a portfolio table (\texttt{has\_portfolio\_table}). This ablation isolates the effect of minimal schema guidance and encourages the model to make its decision transparent through explicit intermediate reasoning. Upon positive detection, extraction is performed identically to the baseline, without additional schema context.

\paragraph{Full Schema Prompting.}
In this ablation, we inject the complete \texttt{Portfolio} Pydantic schema directly into the detection prompt, alongside the page text. The LLM is instructed to reason about the presence of a portfolio table, outputting a chain-of-thought (\texttt{chain\_of\_thought}), a boolean detection (\texttt{has\_portfolio\_table}), and, if present, an extracted \texttt{portfolio} object conforming to the provided schema. This strategy evaluates the effect of strong schema supervision on both detection and extraction performance, requiring the model to both reason and map raw text into the structured schema within a single step.

\paragraph{Direct Schema Application.}
For the final ablation, we bypass explicit table detection and directly apply the \texttt{Portfolio} schema extraction to every page. The LLM is prompted to extract a portfolio table from the provided text and return a Pydantic object with a \texttt{portfolio} field, irrespective of any prior detection or reasoning. Extraction is considered successful if any portion of the schema can be instantiated from the text. This approach evaluates schema-constrained extraction in the absence of explicit detection or intermediate supervision.

\section{Aggregation and Conflict Resolution of Schema Suggestions}
\label{app:aggregation}

After the LLM returns a batch of schema suggestions, we aggregate and cluster similar proposals as follows:

\begin{enumerate}
    \item \textbf{Deduplication}: Suggestions with Levenshtein similarity $\geq 0.9$ (on class name and field structure) are merged.
    \item \textbf{Clustering}: All proposals are clustered by semantic similarity of class names and required fields, using LLMs as the decision process.
    \item \textbf{Selection}: For each cluster, the most frequent or most comprehensive schema suggestion is selected.
    \item \textbf{Validation}: Each selected schema is validated by re-extracting unmatched holdings; suggestions that do not match any holding are dropped.
    \item \textbf{Manual review}: If ambiguity remains, a manual review is triggered for final decision. We validated 64 resolved schemas for the second phase of extraction. Listing~\ref{app:example_recon} displays the reconciled JSON schema for Forward Currency Contract, corresponding Pydantic model via \texttt{pydantic.create\_model}, and several re-extracted holdings.
\end{enumerate}

\section{Schema Suggestion Diversity}
\label{app:diversity}
Table~\ref{tab:diversity} summarizes the diversity among schema suggestions across batch sizes. Moderate batch sizes (100–250) achieve the highest average and maximum diversity, while the largest batch size (500) yields the lowest. This indicates that extremely large batches tend to generate more homogeneous or redundant suggestions, while moderate batches foster a broader range of candidate schemas.

\begin{figure*}[!ht]
    \centering
    \includegraphics[width=0.98\textwidth]{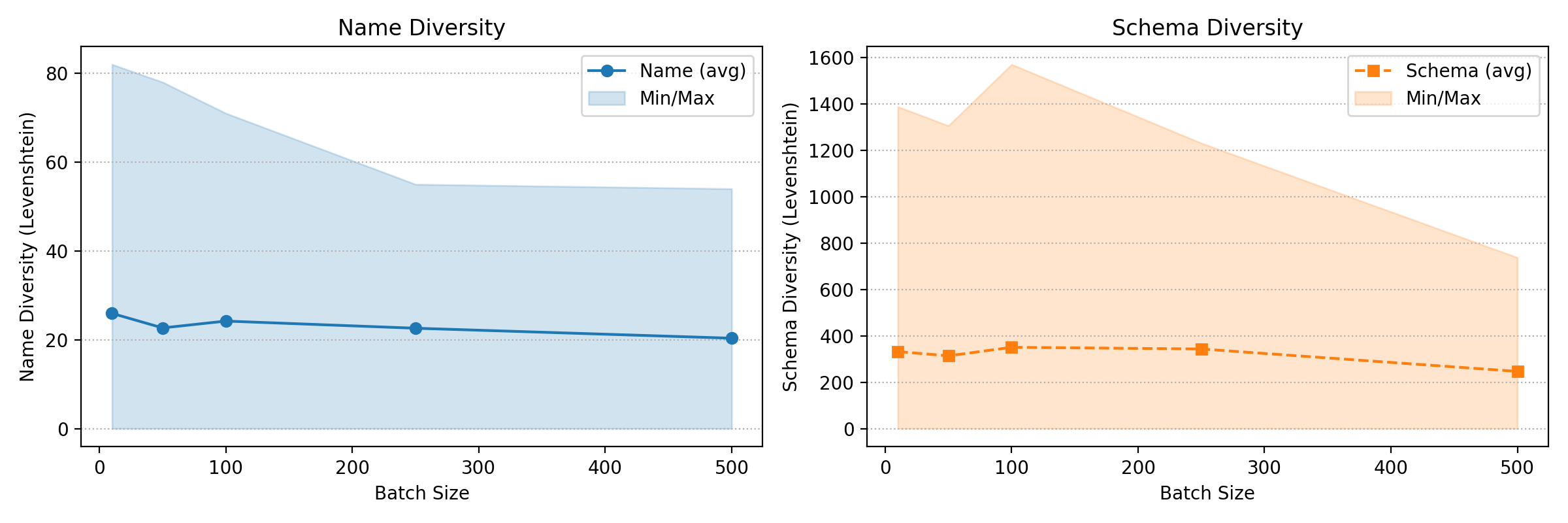}
    \caption{
    \textbf{Left:} Name diversity (average, minimum, and maximum pairwise Levenshtein distance) among schema suggestions for varying batch sizes.
    \textbf{Right:} Schema diversity for the same.
    Moderate batch sizes (100–250) maximize diversity, while very large batches yield more homogeneous outputs.
    }
    \label{fig:diversity-plots}
\end{figure*}

\section{Example Holdings Tables}
\label{app:exampleholdings}
\begin{figure*}
\tiny
\centering
\begin{minipage}{0.5\textwidth}
\centering
\includegraphics[width=0.95\textwidth]{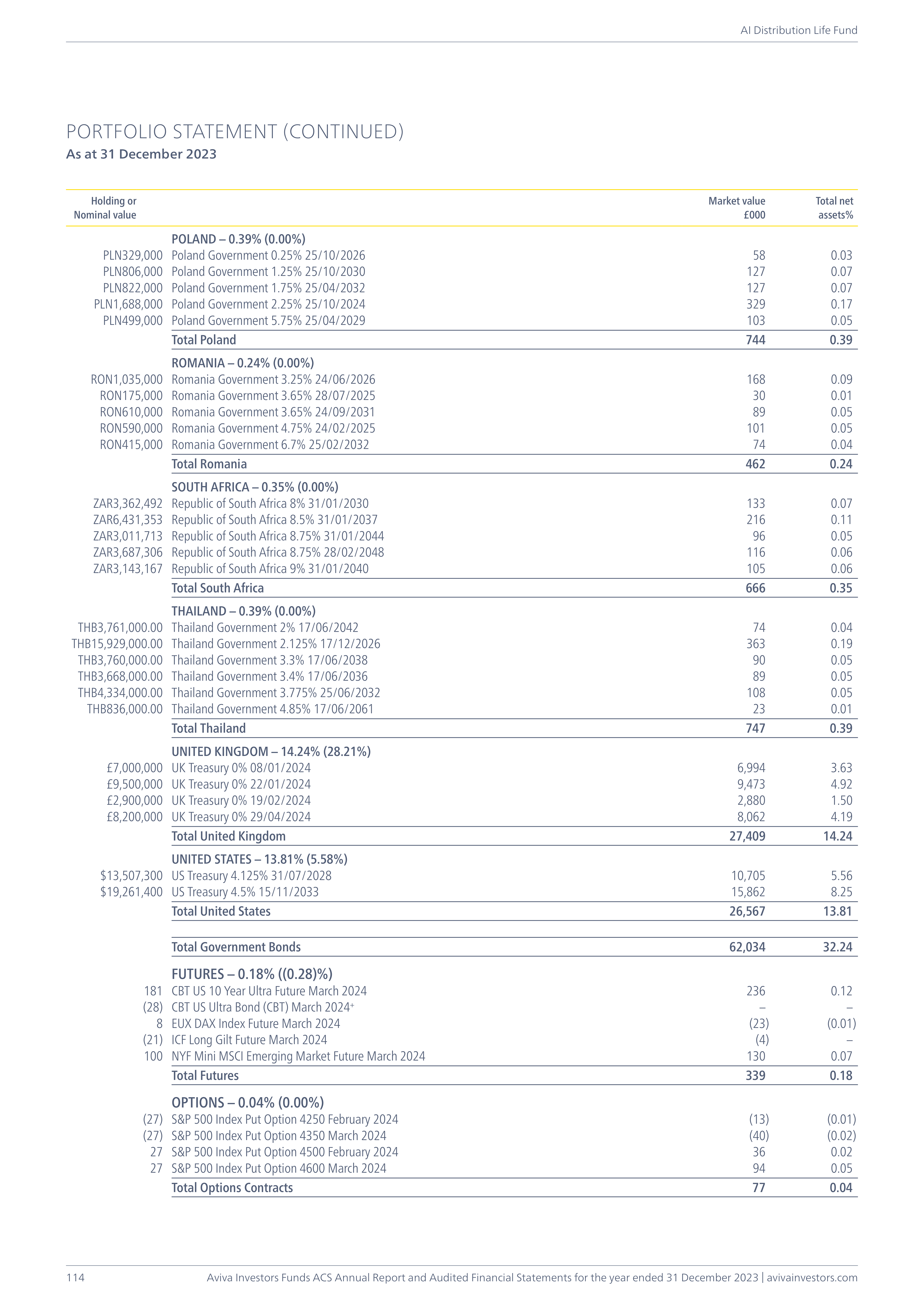}
\caption{Holdings Table Example 1}
\label{fig:Example-image-1}
\end{minipage}\hfill
\centering
\begin{minipage}{0.5\textwidth}
\centering
\resizebox{0.95\textwidth}{!}{%
\begin{tabular}{l|r|r|l|r|c|l|l}
\toprule
{} & {} & \textbf{Market} & {} & \textbf{Coupon} & \textbf{Maturity} &  & \\
\textbf{Description} & \textbf{Quantity} & \textbf{Value} & \textbf{Type} & \textbf{Rate} & \textbf{Date} & \textbf{Issuer} & \textbf{Debt Type} \\
\midrule
Poland Government 0.25\% 25/10/2026 & 329000 & 58000 & Debt & 0.25 & 10/25/2026 & Poland Government & Government Bond \\
Poland Government 1.25\% 25/10/2030 & 806000 & 127000 & Debt & 1.25 & 10/25/2030 & Poland Government & Government Bond \\
Poland Government 1.75\% 25/04/2032 & 822000 & 127000 & Debt & 1.75 & 04/25/2032 & Poland Government & Government Bond \\
Poland Government 2.25\% 25/10/2024 & 1688000 & 329000 & Debt & 2.25 & 10/25/2024 & Poland Government & Government Bond \\
Poland Government 5.75\% 25/04/2029 & 499000 & 103000 & Debt & 5.75 & 04/25/2029 & Poland Government & Government Bond \\
Romania Government 3.25\% 24/06/2026 & 1035000 & 168000 & Debt & 3.25 & 06/24/2026 & Romania Government & Government Bond \\
Romania Government 3.65\% 28/07/2025 & 175000 & 30000 & Debt & 3.65 & 07/28/2025 & Romania Government & Government Bond \\
Romania Government 3.65\% 24/09/2031 & 610000 & 89000 & Debt & 3.65 & 09/24/2031 & Romania Government & Government Bond \\
Romania Government 4.75\% 24/02/2025 & 590000 & 101000 & Debt & 4.75 & 02/24/2025 & Romania Government & Government Bond \\
Romania Government 6.7\% 25/02/2032 & 415000 & 74000 & Debt & 6.7 & 02/25/2032 & Romania Government & Government Bond \\
Republic of South Africa 8\% 31/01/2030 & 3362492 & 133000 & Debt & 8 & 01/31/2030 & Republic of South Africa & Government Bond \\
Republic of South Africa 8.5\% 31/01/2037 & 6431353 & 216000 & Debt & 8.5 & 01/31/2037 & Republic of South Africa & Government Bond \\
Republic of South Africa 8.75\% 31/01/2044 & 3011713 & 96000 & Debt & 8.75 & 01/31/2044 & Republic of South Africa & Government Bond \\
Republic of South Africa 8.75\% 28/02/2048 & 3687306 & 116000 & Debt & 8.75 & 02/28/2048 & Republic of South Africa & Government Bond \\
Republic of South Africa 9\% 31/01/2040 & 3143167 & 105000 & Debt & 9 & 01/31/2040 & Republic of South Africa & Government Bond \\
Thailand Government 2\% 17/06/2042 & 3761000 & 74000 & Debt & 2 & 06/17/2042 & Thailand Government & Government Bond \\
Thailand Government 2.125\% 17/12/2026 & 15929000 & 363000 & Debt & 2.125 & 12/17/2026 & Thailand Government & Government Bond \\
Thailand Government 3.3\% 17/06/2038 & 3760000 & 90000 & Debt & 3.3 & 06/17/2038 & Thailand Government & Government Bond \\
Thailand Government 3.4\% 17/06/2036 & 3668000 & 89000 & Debt & 3.4 & 06/17/2036 & Thailand Government & Government Bond \\
Thailand Government 3.775\% 25/06/2032 & 4334000 & 108000 & Debt & 3.775 & 06/25/2032 & Thailand Government & Government Bond \\
Thailand Government 4.85\% 17/06/2061 & 836000 & 23000 & Debt & 4.85 & 06/17/2061 & Thailand Government & Government Bond \\
UK Treasury 0\% 08/01/2024 & 7000000 & 6994000 & Debt & 0 & 01/08/2024 & UK Treasury & Government Bond \\
UK Treasury 0\% 22/01/2024 & 9500000 & 9473000 & Debt & 0 & 01/22/2024 & UK Treasury & Government Bond \\
UK Treasury 0\% 19/02/2024 & 2900000 & 2880000 & Debt & 0 & 02/19/2024 & UK Treasury & Government Bond \\
UK Treasury 0\% 29/04/2024 & 8200000 & 8062000 & Debt & 0 & 04/29/2024 & UK Treasury & Government Bond \\
US Treasury 4.125\% 31/07/2028 & 13507300 & 10705000 & Debt & 4.125 & 07/31/2028 & US Treasury & Government Bond \\
US Treasury 4.5\% 15/11/2033 & 19261400 & 15862000 & Debt & 4.5 & 11/15/2033 & US Treasury & Government Bond \\
\bottomrule
\end{tabular}
}
\caption{Debt Extracted}

\vspace{1cm}

\centering
\resizebox{0.95\textwidth}{!}{%
\begin{tabular}{l|r|r|c|c}
\toprule
 &  & \textbf{Market} &  & \textbf{Expiration} \\
\textbf{Description} & \textbf{Quantity} & \textbf{Value} & \textbf{Type} & \textbf{Date} \\
\midrule
CBT US 10 Year Ultra Future March 2024 & 181 & 236000 & Future & 03/01/2024 \\
CBT US Ultra Bond (CBT) March 2024+ & -28 & 0 & Future & 03/01/2024 \\
EUX DAX Index Future March 2024 & 8 & -23000 & Future & 03/01/2024 \\
ICF Long Gilt Future March 2024 & -21 & -4000 & Future & 03/01/2024 \\
NYF Mini MSCI Emerging Market Future March 2024 & 100 & 130000 & Future & 03/01/2024 \\
\bottomrule
\end{tabular}
}
\caption{Futures Extracted}

\vspace{1cm}

\centering
\resizebox{0.95\textwidth}{!}{%
\begin{tabular}{l|r|r|l|r|c|c}
\toprule
 &  & \textbf{Market} &  & \textbf{Strike} & \textbf{Expiration} & \textbf{Option} \\
 
\textbf{Description} & \textbf{Quantity} & \textbf{Value} & \textbf{Type} & \textbf{Price} & \textbf{Date} & \textbf{Type} \\
\midrule
S\&P 500 Index Put Option 4250 February 2024 & -27 & -13000 & Option & 4250 & 02/01/2024 & Put \\
S\&P 500 Index Put Option 4350 March 2024 & -27 & -40000 & Option & 4350 & 03/01/2024 & Put \\
S\&P 500 Index Put Option 4500 February 2024 & 27 & 36000 & Option & 4500 & 02/01/2024 & Put \\
S\&P 500 Index Put Option 4600 March 2024 & 27 & 94000 & Option & 4600 & 03/01/2024 & Put \\
\bottomrule
\end{tabular}
}
\caption{Options Extracted}

\end{minipage}
\end{figure*}

\begin{figure*}
\tiny
\centering
\begin{minipage}{0.5\textwidth}
\centering
\includegraphics[width=0.95\textwidth]{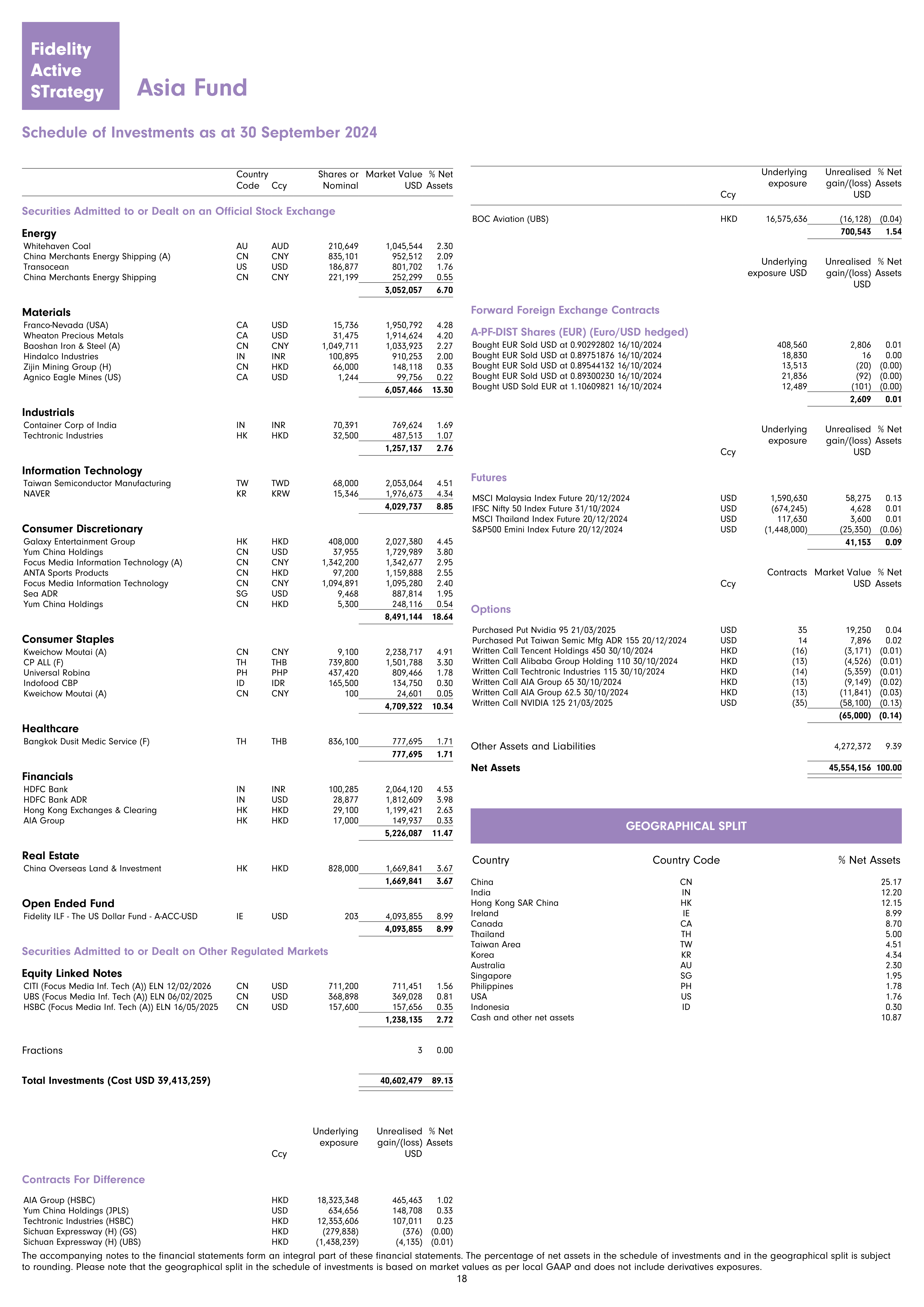}
\caption{Holdings Table Example 2}
\label{fig:Example-image-2}
\end{minipage}\hfill
\centering
\begin{minipage}{0.5\textwidth}
\centering
\resizebox{0.95\textwidth}{!}{%
\begin{tabular}{l|r|r|l|l}
\toprule
 & & \textbf{Market} &  &  \\
\textbf{Description} & \textbf{Quantity} & \textbf{Value} & \textbf{Type} & \textbf{Exch.} \\
\midrule
Whitehaven Coal & 210649 & 1045544 & Equity & AU \\
China Merchants Energy Shipping (A) & 835101 & 952512 & Equity & CN \\
Transocean & 186877 & 801702 & Equity & US \\
China Merchants Energy Shipping & 221199 & 252299 & Equity & CN \\
Franco-Nevada (USA) & 15736 & 1950792 & Equity & CA \\
Wheaton Precious Metals & 31475 & 1914624 & Equity & CA \\
Baoshan Iron \& Steel (A) & 1049711 & 1033923 & Equity & CN \\
Hindalco Industries & 100895 & 910253 & Equity & IN \\
Zijin Mining Group (H) & 66000 & 148118 & Equity & CN \\
Agnico Eagle Mines (US) & 1244 & 99756 & Equity & CA \\
Container Corp of India & 70391 & 769624 & Equity & IN \\
Techtronic Industries & 32500 & 487513 & Equity & HK \\
Taiwan Semiconductor Manufacturing & 68000 & 2053064 & Equity & TW \\
NAVER & 15346 & 1976673 & Equity & KR \\
Galaxy Entertainment Group & 408000 & 2027380 & Equity & HK \\
Yum China Holdings & 37955 & 1729989 & Equity & CN \\
Focus Media Information Technology (A) & 1342200 & 1342677 & Equity & CN \\
ANTA Sports Products & 97200 & 1159888 & Equity & CN \\
Focus Media Information Technology & 1094891 & 1095280 & Equity & CN \\
Sea ADR & 9468 & 887814 & Equity & SG \\
Yum China Holdings & 5300 & 248116 & Equity & CN \\
Kweichow Moutai (A) & 9100 & 2238717 & Equity & CN \\
CP ALL (F) & 739800 & 1501788 & Equity & TH \\
Universal Robina & 437420 & 809466 & Equity & PH \\
Indofood CBP & 165500 & 134750 & Equity & ID \\
Kweichow Moutai (A) & 100 & 24601 & Equity & CN \\
Bangkok Dusit Medic Service (F) & 836100 & 777695 & Equity & TH \\
HDFC Bank & 100285 & 2064120 & Equity & IN \\
HDFC Bank ADR & 28877 & 1812609 & Equity & IN \\
Hong Kong Exchanges \& Clearing & 29100 & 1199421 & Equity & HK \\
AIA Group & 17000 & 149937 & Equity & HK \\
China Overseas Land \& Investment & 828000 & 1669841 & Equity & HK \\
\bottomrule
\end{tabular}
}
\caption{Equities Extracted}

\vspace{0.3cm}

\centering
\resizebox{0.95\textwidth}{!}{%
\begin{tabular}{l|r|r|l|l|l|c}
\toprule
 & & \textbf{Market} & &  & & \textbf{Maturity} \\
\textbf{Description} & \textbf{Quantity} & \textbf{Value} & \textbf{Type} & \textbf{Issuer} & \textbf{Product} & \textbf{Date} \\
\midrule
CITI (Focus Media Inf. Tech (A)) ELN 12/02/2026 & 711200 & 711451 & Equity Linked Note & CITI & Focus Media Inf. Tech (A) & 12/02/2026 \\
UBS (Focus Media Inf. Tech (A)) ELN 06/02/2025 & 368898 & 369028 & Equity Linked Note & UBS & Focus Media Inf. Tech (A) & 02/06/2025 \\
HSBC (Focus Media Inf. Tech (A)) ELN 16/05/2025 & 157600 & 157656 & Equity Linked Note & HSBC & Focus Media Inf. Tech (A) & 05/16/2026 \\
\bottomrule
\end{tabular}
}
\caption{ELNs Extracted}

\vspace{0.3cm}

\centering
\resizebox{0.95\textwidth}{!}{%
\begin{tabular}{l|r|r|l|r|c}
\toprule
& & \textbf{Market} &  & \textbf{Forward} & \textbf{Settlement} \\
\textbf{Description} & \textbf{Quantity} & \textbf{Value} & \textbf{Type} & \textbf{Price} & \textbf{Date} \\
\midrule
Bought EUR Sold USD at 0.90292802 16/10/2024 & 408560 & 2806 & Forward & 0.90292802 & 10/16/2024 \\
Bought EUR Sold USD at 0.89751876 16/10/2024 & 18830 & 16 & Forward & 0.89751876 & 10/16/2024 \\
Bought EUR Sold USD at 0.89544132 16/10/2024 & 13513 & -20 & Forward & 0.89544132 & 10/16/2024 \\
Bought EUR Sold USD at 0.89300230 16/10/2024 & 21836 & -92 & Forward & 0.89300230 & 10/16/2024 \\
Bought USD Sold EUR at 1.10609821 16/10/2024 & 12489 & -101 & Forward & 1.10609821 & 10/16/2024 \\
\bottomrule
\end{tabular}
}
\caption{Forwards Extracted}

\vspace{0.3cm}

\centering
\resizebox{0.95\textwidth}{!}{%
\begin{tabular}{l|r|r|l|r|c}
\toprule
 &  & \textbf{Market} &  & \textbf{Contract} & \textbf{Expiration} \\
\textbf{Description} & \textbf{Quantity} & \textbf{Value} & \textbf{Type} & \textbf{Size} & \textbf{Date} \\
\midrule
MSCI Malaysia Index Future 20/12/2024 & 1590630 & 58275 & Future & 1590630 & 12/20/2024 \\
IFSC Nifty 50 Index Future 31/10/2024 & -674245 & 4628 & Future & 674245 & 10/31/2024 \\
MSCI Thailand Index Future 20/12/2024 & 117630 & 3600 & Future & 117630 & 12/20/2024 \\
S\&P500 Emini Index Future 20/12/2024 & -1448000 & -25350 & Future & 1448000 & 12/20/2024 \\
\bottomrule
\end{tabular}
}
\caption{Futures Extracted}

\vspace{0.3cm}

\centering
\resizebox{0.95\textwidth}{!}{%
\begin{tabular}{l|r|r|c|l|r|c|c|l}
\toprule
& & \textbf{Market} &  &  & \textbf{Strike} & \textbf{Expiration} & \textbf{Option} &  \\
\textbf{Description} & \textbf{Quantity} & \textbf{Value} & \textbf{Type} & \textbf{Underlying} & \textbf{Price} & \textbf{Date} & \textbf{Type} & \textbf{Ticker} \\
\midrule
Purchased Put Nvidia 95 21/03/2025 & 35 & 19250 & Option & Nvidia & 95 & 03/21/2025 & Put & Nvidia \\
Purchased Put Taiwan Semic Mfg ADR 155 20/12/2024 & 14 & 7896 & Option & Taiwan Semic Mfg ADR & 155 & 12/20/2024 & Put & Taiwan Semic Mfg ADR \\
Written Call Tencent Holdings 450 30/10/2024 & -16 & -3171 & Option & Tencent Holdings & 450 & 10/30/2024 & Call & Tencent Holdings \\
Written Call Alibaba Group Holding 110 30/10/2024 & -13 & -4526 & Option & Alibaba Group Holding & 110 & 10/30/2024 & Call & Alibaba Group Holding \\
Written Call Techtronic Industries 115 30/10/2024 & -14 & -5359 & Option & Techtronic Industries & 115 & 10/30/2024 & Call & Techtronic Industries \\
Written Call AIA Group 65 30/10/2024 & -13 & -9149 & Option & AIA Group & 65 & 10/30/2024 & Call & AIA Group \\
Written Call AIA Group 62.5 30/10/2024 & -13 & -11841 & Option & AIA Group & 62.5 & 10/30/2024 & Call & AIA Group \\
Written Call NVIDIA 125 21/03/2025 & -35 & -58100 & Option & NVIDIA & 125 & 03/21/2025 & Call & NVIDIA \\
\bottomrule
\end{tabular}
}
\caption{Options Extracted}

\end{minipage}
\end{figure*}

\begin{figure*}
\tiny
\centering
\begin{minipage}{0.55\textwidth}
\centering
\includegraphics[width=0.95\textwidth]{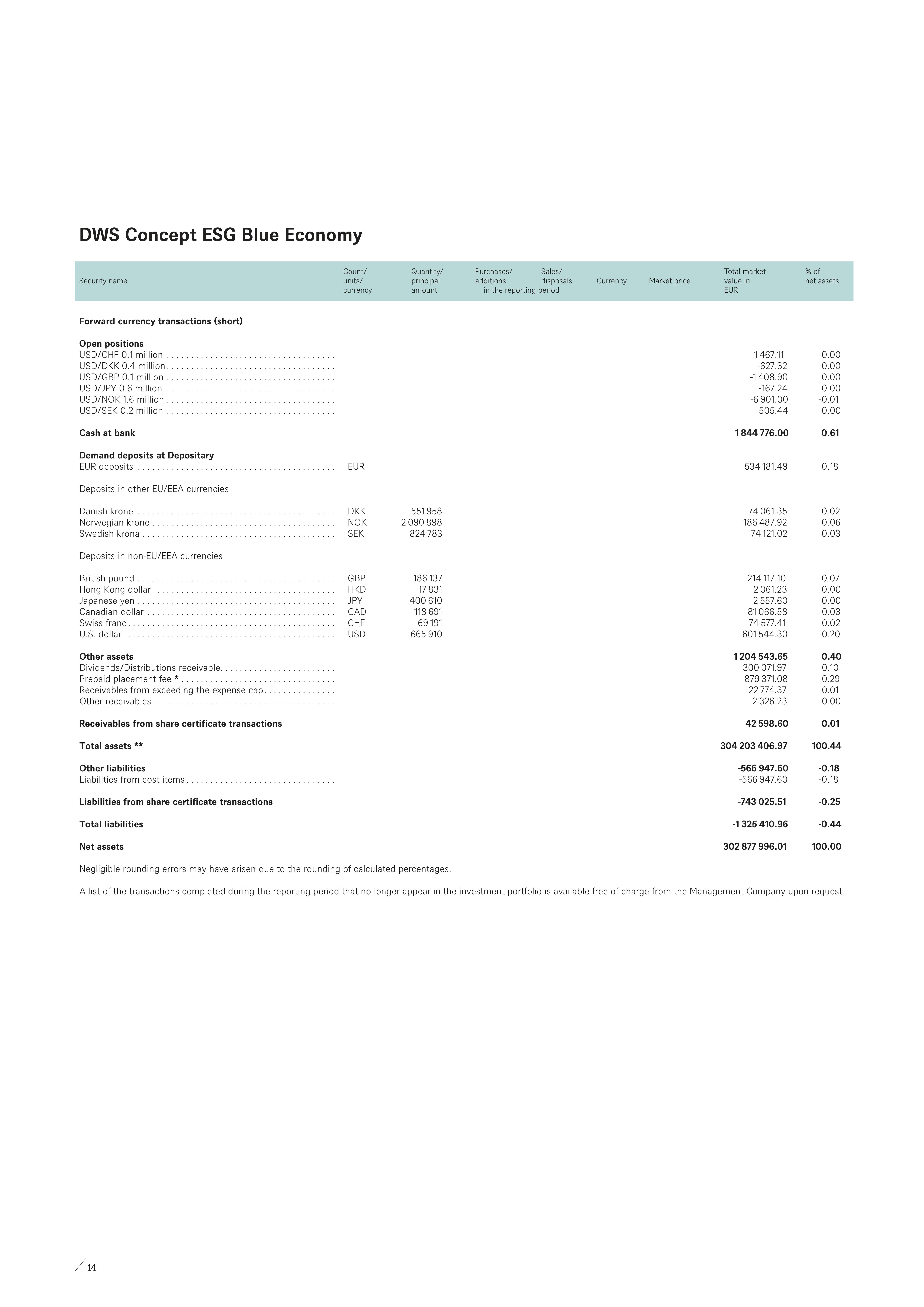}
\caption{Holdings Table Example 3}
\label{fig:Example-image-3}
\end{minipage}\hfill
\begin{minipage}{0.45\textwidth}
\centering
\resizebox{\linewidth}{!}{%
\begin{tabular}{l|r|c}
\toprule
 & \textbf{Market} &  \\
\textbf{Description} & \textbf{Value} & \textbf{Type} \\
\midrule
USD/CHF 0.1 million & -1467.11 & Forward \\
USD/DKK 0.4 million & -627.32 & Forward \\
USD/GBP 0.1 million & -1408.9 & Forward \\
USD/JPY 0.6 million & -167.24 & Forward \\
USD/NOK 1.6 million & -6901 & Forward \\
USD/SEK 0.2 million & -505.44 & Forward \\
\bottomrule
\end{tabular}
}
\caption{Forwards Extracted}

\vspace{1cm}

\centering
\resizebox{\textwidth}{!}{%
\begin{tabular}{l|c|r}
\toprule
& & \textbf{Market} \\
\textbf{Description} & \textbf{Type} & \textbf{Value} \\
\midrule
Cash at bank & Other & 1844776 \\
Demand deposits at Depositary - EUR deposits & Other & 534181.49 \\
Deposits in other EU/EEA currencies - Danish krone & Other & 74061.35 \\
Deposits in other EU/EEA currencies - Norwegian krone & Other & 186487.92 \\
Deposits in other EU/EEA currencies - Swedish krona & Other & 74121.02 \\
Deposits in non-EU/EEA currencies - British pound & Other & 214117.1 \\
Deposits in non-EU/EEA currencies - Hong Kong dollar & Other & 2061.23 \\
Deposits in non-EU/EEA currencies - Japanese yen & Other & 2557.6 \\
Deposits in non-EU/EEA currencies - Canadian dollar & Other & 81066.58 \\
Deposits in non-EU/EEA currencies - Swiss franc & Other & 74577.41 \\
Deposits in non-EU/EEA currencies - U.S. dollar & Other & 601544.3 \\
Dividends/Distributions receivable & Other & 300071.97 \\
Prepaid placement fee & Other & 879371.08 \\
Receivables from exceeding the expense cap & Other & 22774.37 \\
Other receivables & Other & 2326.23 \\
Receivables from share certificate transactions & Other & 42598.6 \\
Liabilities from cost items & Other & -566947.6 \\
Liabilities from share certificate transactions & Other & -743025.51 \\
\bottomrule
\end{tabular}
}
\caption{Other Instruments Extracted}

\end{minipage}
\end{figure*}

\begin{figure*}
\tiny
\centering
\begin{minipage}{0.5\textwidth}
\centering
\includegraphics[width=\linewidth]{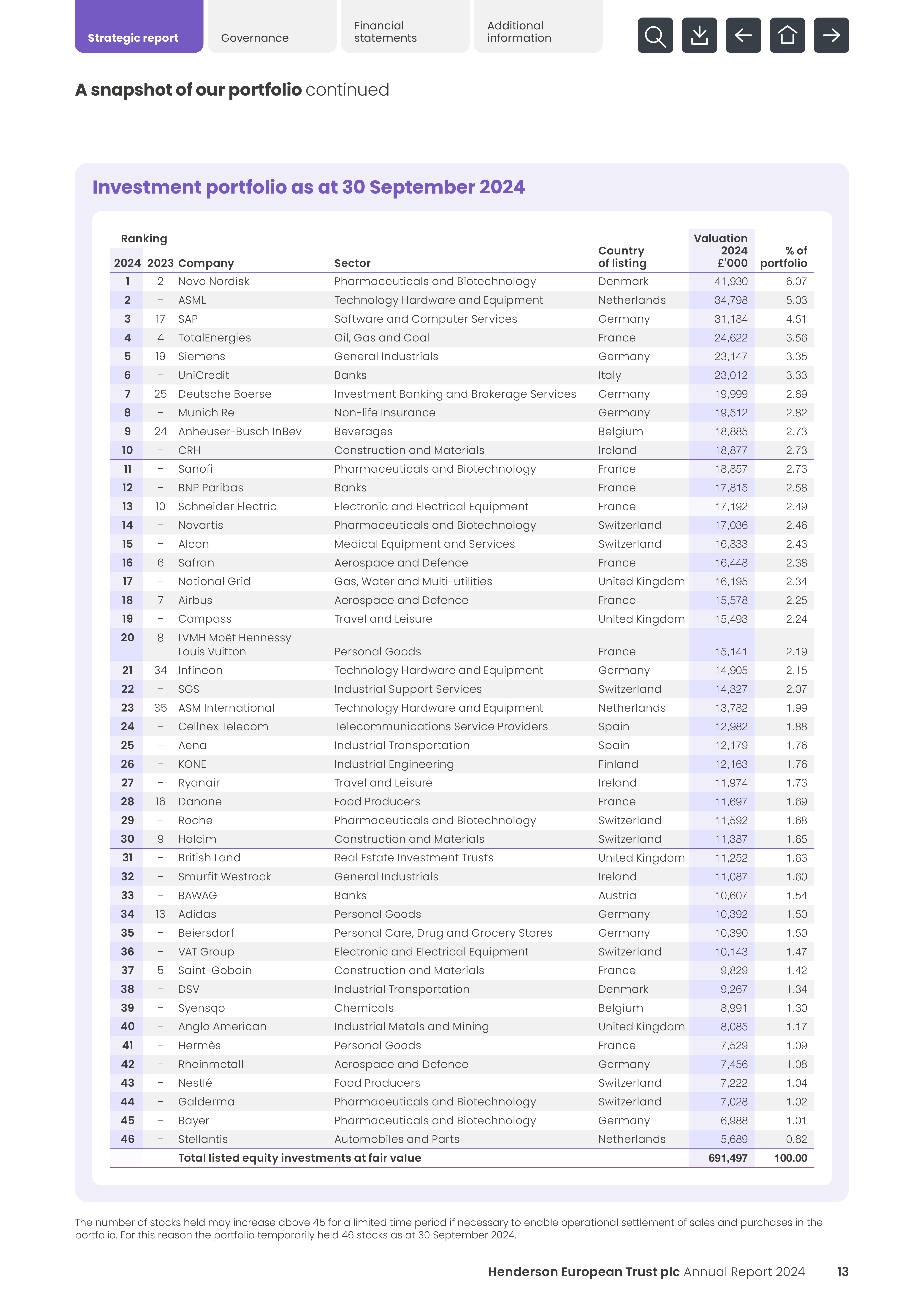}
\caption{Holdings Table Example 4}
\label{fig:Example-image-4}
\end{minipage}\hfill
\begin{minipage}{0.45\textwidth}
\centering
\resizebox{\linewidth}{!}{%
\begin{tabular}{l|r|c}
\toprule
& \textbf{Market} &  \\
\textbf{Description} & \textbf{Value} & \textbf{Type} \\
\midrule
Novo Nordisk & 41930000 & Equity \\
ASML & 34798000 & Equity \\
SAP & 31184000 & Equity \\
TotalEnergies & 24622000 & Equity \\
Siemens & 23147000 & Equity \\
UniCredit & 23012000 & Equity \\
Deutsche Boerse & 19999000 & Equity \\
Munich Re & 19512000 & Equity \\
Anheuser-Busch InBev & 18885000 & Equity \\
CRH & 18877000 & Equity \\
Sanofi & 18857000 & Equity \\
BNP Paribas & 17815000 & Equity \\
Schneider Electric & 17192000 & Equity \\
Novartis & 17036000 & Equity \\
Alcon & 16833000 & Equity \\
Safran & 16448000 & Equity \\
National Grid & 16195000 & Equity \\
Airbus & 15578000 & Equity \\
Compass & 15493000 & Equity \\
LVMH Moët Hennessy Louis Vuitton & 15141000 & Equity \\
Infineon & 14905000 & Equity \\
SGS & 14327000 & Equity \\
ASM International & 13782000 & Equity \\
Cellnex Telecom & 12982000 & Equity \\
Aena & 12179000 & Equity \\
KONE & 12163000 & Equity \\
Ryanair & 11974000 & Equity \\
Danone & 11697000 & Equity \\
Roche & 11592000 & Equity \\
Holcim & 11387000 & Equity \\
British Land & 11252000 & Equity \\
Smurfit Westrock & 11087000 & Equity \\
BAWAG & 10607000 & Equity \\
Adidas & 10392000 & Equity \\
Beiersdorf & 10390000 & Equity \\
VAT Group & 10143000 & Equity \\
Saint-Gobain & 9829000 & Equity \\
DSV & 9267000 & Equity \\
Syensqo & 8991000 & Equity \\
Anglo American & 8085000 & Equity \\
Hermès & 7529000 & Equity \\
Rheinmetall & 7456000 & Equity \\
Nestlé & 7222000 & Equity \\
Galderma & 7028000 & Equity \\
Bayer & 6988000 & Equity \\
Stellantis & 5689000 & Equity \\
\bottomrule
\end{tabular}
}
\caption{Equities Extracted}
\end{minipage}
\end{figure*}

\begin{figure*}[!ht]
\tiny
\centering
\begin{minipage}{0.5\textwidth}
\centering
\includegraphics[width=\textwidth]{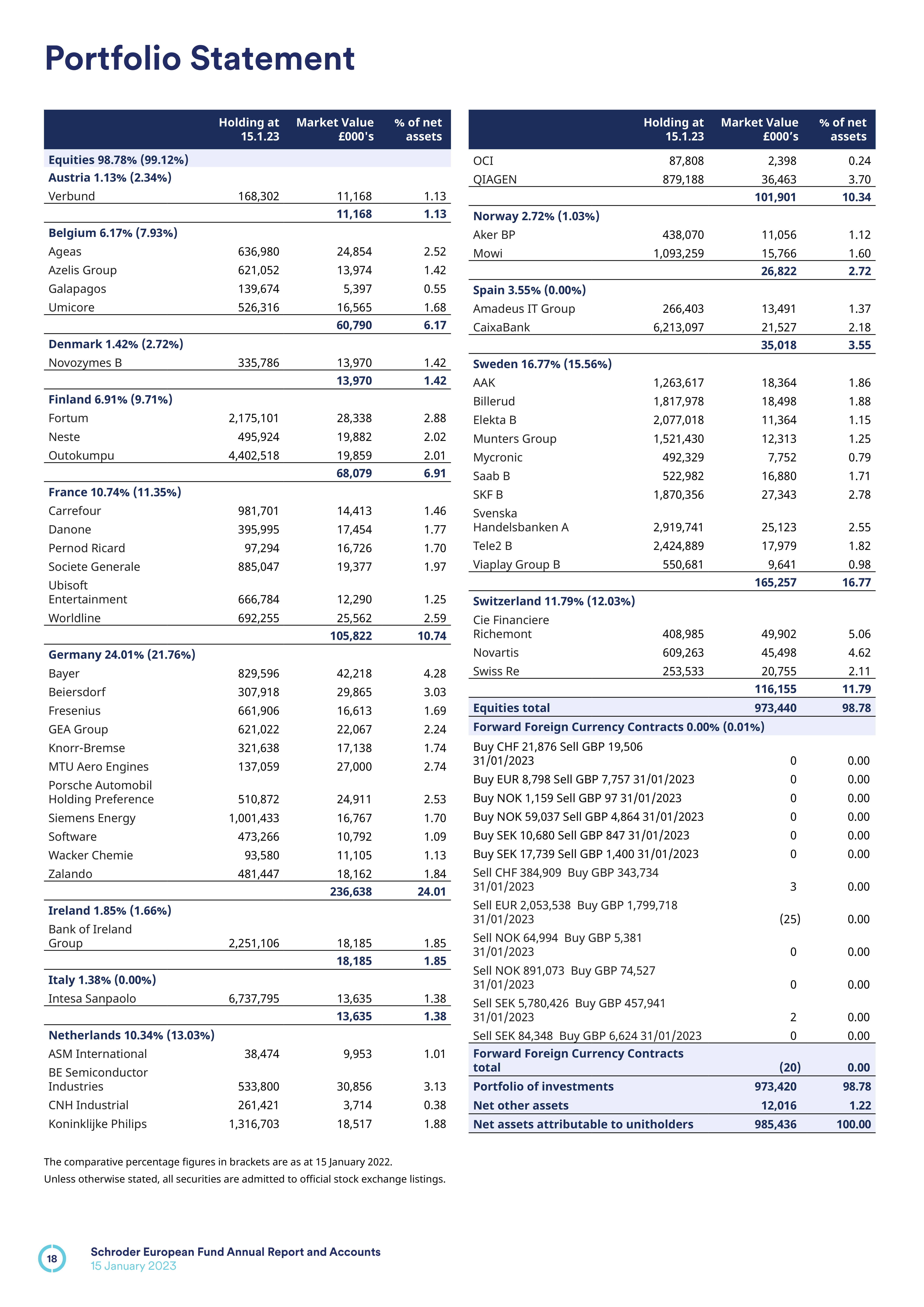}
\caption{Holdings Table Example 5}
\label{fig:Example-image-5}
\end{minipage}\hfill
\begin{minipage}{0.48\textwidth}
\centering
\resizebox{\textwidth}{!}{%
\begin{tabular}{l|r|r|l}
\toprule
\textbf{Description} & \textbf{Quantity} & \textbf{Market Value} & \textbf{Type} \\
\midrule
Verbund & 168302 & 11168000 & Equity \\
Ageas & 636980 & 24854000 & Equity \\
Azelis Group & 621052 & 13974000 & Equity \\
Galapagos & 139674 & 5397000 & Equity \\
Umicore & 526316 & 16565000 & Equity \\
Novozymes B & 335786 & 13970000 & Equity \\
Fortum & 2175101 & 28338000 & Equity \\
Neste & 495924 & 19882000 & Equity \\
Outokumpu & 4402518 & 19859000 & Equity \\
Carrefour & 981701 & 14413000 & Equity \\
Danone & 395995 & 17454000 & Equity \\
Pernod Ricard & 97294 & 16726000 & Equity \\
Societe Generale & 885047 & 19377000 & Equity \\
Ubisoft Entertainment & 666784 & 12290000 & Equity \\
Worldline & 692255 & 25562000 & Equity \\
Bayer & 829596 & 42218000 & Equity \\
Beiersdorf & 307918 & 29865000 & Equity \\
Fresenius & 661906 & 16613000 & Equity \\
GEA Group & 621022 & 22067000 & Equity \\
Knorr-Bremse & 321638 & 17138000 & Equity \\
MTU Aero Engines & 137059 & 27000000 & Equity \\
Porsche Automobil Holding Preference & 510872 & 24911000 & Equity \\
Siemens Energy & 1001433 & 16767000 & Equity \\
Software & 473266 & 10792000 & Equity \\
Wacker Chemie & 93580 & 11105000 & Equity \\
Zalando & 481447 & 18162000 & Equity \\
Bank of Ireland Group & 2251106 & 18185000 & Equity \\
Intesa Sanpaolo & 6737795 & 13635000 & Equity \\
ASM International & 38474 & 9953000 & Equity \\
BE Semiconductor Industries & 533800 & 30856000 & Equity \\
CNH Industrial & 261421 & 3714000 & Equity \\
Koninklijke Philips & 1316703 & 18517000 & Equity \\
OCI & 87808 & 2398000 & Equity \\
QIAGEN & 879188 & 36463000 & Equity \\
Aker BP & 438070 & 11056000 & Equity \\
Mowi & 1093259 & 15766000 & Equity \\
Amadeus IT Group & 266403 & 13491000 & Equity \\
CaixaBank & 6213097 & 21527000 & Equity \\
AAK & 1263617 & 18364000 & Equity \\
Billerud & 1817978 & 18498000 & Equity \\
Elekta B & 2077018 & 11364000 & Equity \\
Munters Group & 1521430 & 12313000 & Equity \\
Mycronic & 492329 & 7752000 & Equity \\
Saab B & 522982 & 16880000 & Equity \\
SKF B & 1870356 & 27343000 & Equity \\
Svenska Handelsbanken A & 2919741 & 25123000 & Equity \\
Tele2 B & 2424889 & 17979000 & Equity \\
Viaplay Group B & 550681 & 9641000 & Equity \\
Cie Financiere Richemont & 408985 & 49902000 & Equity \\
Novartis & 609263 & 45498000 & Equity \\
Swiss Re & 253533 & 20755000 & Equity \\
\bottomrule
\end{tabular}
}
\caption{Equities Extracted}

\vspace{1cm}

\centering
\resizebox{\textwidth}{!}{%
\begin{tabular}{l|l|r}
\toprule
\textbf{Description} & \textbf{Type} & \textbf{Market Value} \\
\midrule
Net other assets & Other & 12016000 \\
\bottomrule
\end{tabular}
}
\caption{Other Instruments Extracted}
\end{minipage}
\end{figure*}

\begin{figure*}[!ht]
\tiny
\centering
\begin{minipage}{0.5\textwidth}
\centering
\includegraphics[width=\textwidth]{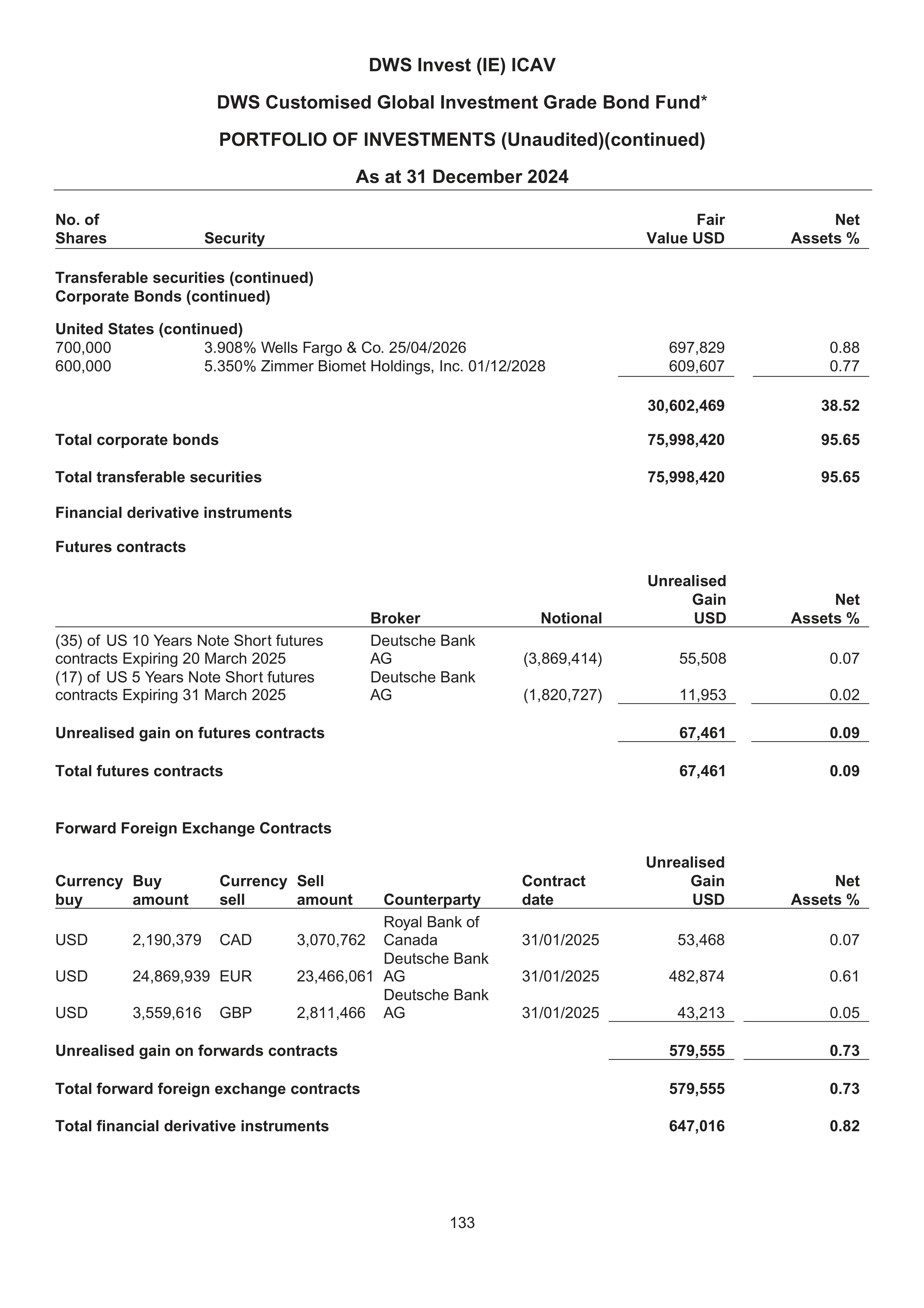}
\caption{Holdings Table Example 6}
\label{fig:Example-image-6}
\end{minipage}\hfill
\begin{minipage}{0.48\textwidth}
\centering
\resizebox{\textwidth}{!}{%
\begin{tabular}{l|l|l|c|l|c|l}
\toprule
& & \textbf{Market} &  & \textbf{Coupon} & \textbf{Maturity} &  \\
\textbf{Description} & \textbf{Quantity} & \textbf{Value} & \textbf{Type} & \textbf{Rate} & \textbf{Date} & \textbf{Issuer} \\
\midrule
CBT US 10 Year Ultra Future March 2024 & 181 & 236000 & Future & 4250 & 02/01/2024 & Poland Government \\
CBT US Ultra Bond (CBT) March 2024+ & -28 & 0 & Future & 4250 & 03/01/2024 & Poland Government \\
EUX DAX Index Future March 2024 & 8 & -23000 & Future & 4350 & 03/01/2024 & \\
ICF Long Gilt Future March 2024 & -21 & -4000 & Future & 4500 & 02/01/2024 & \\
NYF Mini MSCI Emerging Market Future March 2024 & 100 & 130000 & Future & 4600 & 03/01/2024 & \\
\bottomrule
\end{tabular}
}
\caption{Debt Extracted}

\vspace{1cm}

\centering
\resizebox{\textwidth}{!}{%
\begin{tabular}{l|r|r|l|l}
\toprule
 &  & \textbf{Market} &  & \textbf{Expiration} \\
\textbf{Description} & \textbf{Quantity} & \textbf{Value} & \textbf{Type} & \textbf{Date} \\
\midrule
US 10 Years Note Short futures contracts & -35 & 55508 & Future & 02/03/2025 \\
US 5 Years Note Short futures contracts & -17 & 11953 & Future & 31/03/2025 \\
\bottomrule
\end{tabular}
}
\caption{Futures Extracted}

\vspace{1cm}

\centering
\resizebox{\textwidth}{!}{%
\begin{tabular}{l|r|r|c|c}
\toprule
& & \textbf{Market} & \textbf{Type} & \textbf{Settlement} \\
\textbf{Description} & \textbf{Quantity} & \textbf{Value} & \textbf{Type} & \textbf{Date} \\
\midrule
USD/CAD Forward Contract & 1 & 53468 & Forward & 31/01/2025 \\
USD/EUR Forward Contract & 1 & 482874 & Forward & 31/01/2025 \\
USD/GBP Forward Contract & 1 & 43213 & Forward & 31/01/2025 \\
\bottomrule
\end{tabular}
}
\caption{Forwards Extracted}

\end{minipage}
\end{figure*}

We show example holdings tables, alongside TASER's extractions in Figures~\ref{fig:Example-image-1}~-~\ref{fig:Example-image-6}.

\begin{table*}[!ht]
\centering

\begin{tabular}{c|ccc|ccc|cc}
\toprule
\small
\textbf{Batch} & \multicolumn{3}{c|}{\textbf{Coverage (Holdings)}} & \multicolumn{3}{c|}{\textbf{Utilization (Schemas)}} & \textbf{Reported} & \textbf{Collision} \\
\textbf{Size} & \textbf{Count} & \textbf{\% Covered} & \textbf{NAV (\%)} &  \textbf{Total \#} & \textbf{\# Utilized} & \textbf{\%} & \textbf{Collisions} & \textbf{Rate (\%)}\\
\midrule
10  & 16,942 & 96.1 & 99.65 & 867 & 251 & 29.0 & 2,409 & 73.5\\
50   & 7,416 & 42.1 & 35.35 & 586 & 240 & 41.0 & 442 & 57.0\\
100  & 7,311 & 41.5 & 35.39 & 495 & 217 & 43.8 & 218 & 30.6 \\
250  & 6,955 & 39.5 & 35.46 & 351 & 156 & 44.4 & 75 & 17.6 \\
500  & 6,349 & 36.0 & 35.10 & 184 & 109 & 59.2 & 30 & 14.0 \\
\bottomrule
\end{tabular}
\caption{\textbf{Schema Utilization Efficiency.}
We report the proportion of generated schema suggestions that were utilized (i.e., matched at least one holding), for matching. Larger batch sizes result in a higher fraction of utilized schemas, suggesting that bulkier suggestion rounds are more efficient at targeting actionable schemas, albeit at the expense of overall diversity and coverage.}
\label{tab:schema-utilization}
\end{table*}

\begin{table*}[!ht]
\small
\centering
\resizebox{\linewidth}{!}{
\begin{tabular}{clcccccc}
\toprule
& {} & \multicolumn{2}{c}{\textbf{Detection}} & \multicolumn{2}{c}{\textbf{Extraction}} & \multicolumn{2}{c}{\textbf{End to End}}\\
\cmidrule(lr){3-4} \cmidrule(lr){5-6} \cmidrule(lr){7-8}
& \textbf{Method}
& \textbf{Tokens} & \textbf{Latency (s)}
                & \textbf{Tokens} & \textbf{Latency (s)}
                & \textbf{Tokens} & \textbf{Latency (s)} \\
\midrule
\multirow{4}{*}{\rotatebox{90}{\textbf{TASER}}} & 
(a) Raw Text Prompting        & 1,495 & 0.33 & 5,414 & 20.37 & 6,909  & 20.69\\
& (b) Structured CoT            & 1,514 & 1.70 & 5,440 & 20.20 & 6,954  & 21.90\\
& (c) Full Schema Prompting     & 5,706 & 1.58 & 5,235 & 20.48 & 10,941 & 22.07\\
& (d) Direct Schema Application & ---   & ---  & 5,693 & 21.47 & 5,693  & 21.47 \\
\bottomrule
\end{tabular}
}
\caption{\textbf{Efficiency comparison of each ablation strategy}. We report the token consumption and inference latency for detection, extraction, and end-to-end processing. Raw Text Prompting minimizes detection cost (1,495 tokens, 0.33 s) and achieves a total pipeline latency of 20.69 s; Structured CoT incurs additional reasoning overhead (1.70 s) with similar extraction performance; Full Schema Prompting uses the most detection tokens (5,706) but maintains comparable end-to-end latency (22.07 s); Direct Schema Application skips the detection stage entirely, applying schema validation directly in extraction. Dashes (---) indicate stages not performed by the method.}
\label{tab:efficiency}
\end{table*}

\begin{table*}[t]
\centering
\begin{tabular}{lrl}
\toprule
\textbf{Instrument Category}    & \textbf{Count}   & \textbf{Example} \\
\midrule
Equities              & 28,737  & Taiwan Semiconductor Manufacturing \\
Debt                  & 17,105  & US Treasury 4.69\% 09/05/2024 \\
Unmatched Instruments      & 16,822  & EUR \\
Forwards              & 8,023   & Bought USD Sold KRW at 0.00072513 \\
Options               &   977   & Written Call Unilever 4050 \\
Futures               &   720   & US 5 Year Bond Future \\
Swaps                 &   776   & Pay fixed 3.026\% receive float. (1d SOFR) \\
ELNs                  &   292   & BNP (Laobaixing Pharm. Chain (A)) ELN 22/07/2024 \\
\bottomrule
\end{tabular}
\caption{\textbf{Distribution of instrument categories} in the dataset, with an example for each.}
\label{tab:instrument-categories}
\end{table*}

\begin{figure*}
\begin{lstlisting}[language=Python,numbers=left,
  numbersep=6pt,
  xleftmargin=2.2em]
# Ablation 1: Raw Text Prompting
detection_prompt = (
    "Is there a table present in the following text? Reply with 'yes' or 'no'.\n\n"
    f"Text:\n{page.text}"
)


# TableDetectionResponse Pydantic Model
class TableDetectionResponse(BaseModel):
    table_chain_of_thought: str = Field(..., 
        description="Chain of thoughts on if the page text contains table-like content")
    has_portfolio_table: bool = Field(..., 
        description="True if the page has a holdings table, False otherwise")


# Ablation 2: Structured Chain-of-Thought (CoT)
detection_prompt = (
    "Analyze the following text and determine if it contains a portfolio table. "
    "Provide your chain of thought and final decision in a structured output "
    "response model that includes 'chain_of_thought' and 'has_portfolio_table' fields.\n\n"
    f"Text:\n{page.text}"
)


# Ablation 3: Full Schema Prompting
detection_prompt = (
    "Using the provided Portfolio JSON schema, analyze the following text and "
    "if it can be extracted into that schema. Provide your chain of thought. "
    "You will output a response model object including 'chain_of_thought', "
    "'has_portfolio_table', and 'extracted portfolio'.\n\n"
    f"Schema:\n{json.dumps(schema, indent=2)}\n\n"
    f"Text:\n{page.text}"
)


# Ablation 4: Direct Schema Application
detection_prompt = (
    "Extract a portfolio table from the following text following the Portfolio schema. "
    "Return a response object with a 'portfolio' field.\n\n"
    f"Text:\n{page.text}"
)
\end{lstlisting}
\caption{\textbf{Detection prompts for all ablation strategies.} Each section is labeled with its corresponding ablation strategy.} \label{figure:detection prompt}
\end{figure*}



\begin{figure*}[ht]
\begin{lstlisting}[language=Python, caption={Portfolio schema with all matched instrument types.},numbers=left,
  numbersep=6pt,
  xleftmargin=2.2em, breaklines=true,
  breakatwhitespace=true,
  columns=fullflexible,
  keepspaces=true]
from enum import Enum
from typing import Optional, List, Literal
from pydantic import BaseModel, Field
from datetime import datetime

class BaseInstrument(BaseModel):
    cusip: Optional[str] = Field(None, description="CUSIP identifier")
    isin: Optional[str] = Field(None, 
        description="International Securities Identification Number")
    ticker: Optional[str] = Field(None, description="Ticker Symbol")
    description: Optional[str] = Field(None, 
        description="Description or name of the instrument")
    quantity: Optional[float] = Field(None, description="Number of units held")
    market_value: Optional[float] = Field(None, description="Market value of the holding")

class Equity(BaseInstrument):
    instrument_type: Literal["Equity"] = "Equity"
    exchange: Optional[str] = Field(None, description="Trading exchange for the equity")

class Option(BaseInstrument):
    instrument_type: Literal["Option"] = "Option"
    underlying: Optional[str] = Field(None, description="Identifier for the underlying asset")
    strike_price: Optional[float] = Field(None, description="Strike price of the option")
    expiration_date: Optional[datetime] = Field(None, 
        description="Expiration date of the option")
    option_type: Optional[str] = Field(None, description="Call or Put option")

class Swap(BaseInstrument):
    instrument_type: Literal["Swap"] = "Swap"
    notional_amount: Optional[float] = Field(None, description="Notional amount of the swap")
    fixed_rate: Optional[float] = Field(None, 
        description="Fixed rate component (if applicable)")
    floating_rate_index: Optional[str] = Field(None, 
        description="Index used for floating rate leg")
    maturity_date: Optional[datetime] = Field(None, description="Maturity date of the swap")
    counterparty: Optional[str] = Field(None, description="The name of the counterparty")

class Forward(BaseInstrument):
    instrument_type: Literal["Forward"] = "Forward"
    forward_price: Optional[float] = Field(None, description="Agreed forward price")
    settlement_date: Optional[datetime] = Field(None, 
        description="Settlement date for the forward")

class Future(BaseInstrument):
    instrument_type: Literal["Future"] = "Future"
    contract_size: Optional[int] = Field(None, description="Size of the contract")
    expiration_date: Optional[datetime] = Field(None, 
        description="Expiration date of the future")

class Debt(BaseInstrument):
    instrument_type: Literal["Debt"] = "Debt"
    coupon_rate: Optional[float] = Field(None, 
        description="Annual coupon rate of the debt/bond")
    maturity_date: Optional[datetime] = Field(None, 
        description="Maturity date of the debt/bond")
    issuer: Optional[str] = Field(None, description="Issuer of the debt/bond")

class EquityLinkedNote(BaseInstrument):
    instrument_type: Literal["Equity Linked Note"] = "Equity Linked Note"
    issuer: Optional[str] = Field(None, description="Issuer of the ELN")
    product: Optional[str] = Field(None, description="Underlying product of the ELN")
    maturity_date: Optional[datetime] = Field(None, description="Maturity date of the ELN")
\end{lstlisting}
\end{figure*}

\begin{figure*}
    \begin{lstlisting}[language=Python, caption={Main Portfolio Model with Unmatched (Other) Holdings class},numbers=left,
  numbersep=6pt,
  xleftmargin=2.2em, breaklines=true,
  breakatwhitespace=true,
  columns=fullflexible,
  keepspaces=true]
class Other(BaseModel):
    description: str = Field(..., 
        description="Text of the unknown instrument.")
    name: str = Field(..., 
        description="Suggested classification of the description or type")
    market_value: Optional[float] = Field(None, 
        description="Market value associated with the instrument"
    )
    
class Portfolio(BaseModel):
    fund_name: Optional[str] = Field(None, 
        description="Name of the fund that the portfolio belongs to")
    value_in_thousands: bool = Field(False, 
        description="True if the market value is based on thousands")
    equities: Optional[List[Equity]] = Field(default_factory=list, 
        description="List of equities")
    options: Optional[List[Option]] = Field(default_factory=list, 
        description="List of options")
    swaps: Optional[List[Swap]] = Field(default_factory=list, 
        description="List of swaps")
    forwards: Optional[List[Forward]] = Field(default_factory=list, 
        description="List of forwards")
    futures: Optional[List[Future]] = Field(default_factory=list, 
        description="List of futures")
    debt: Optional[List[Debt]] = Field(default_factory=list, 
        description="List of debt instruments")
    elns: Optional[List[EquityLinkedNote]] = Field(default_factory=list, 
        description="List of equity linked notes")
    other_instruments: Optional[List[Other]] = Field(default_factory=list, 
        description="The list of instruments that do not match any other type")
    \end{lstlisting}
\end{figure*}

\begin{figure*}
\begin{lstlisting}[language=json,numbers=left,
  numbersep=6pt,
  xleftmargin=2.2em, breaklines=true,
  breakatwhitespace=true,
  columns=fullflexible,
  keepspaces=true]
# Prompt template for Recommender Agent, using batch size parameterization

def recommender_agent_prompt(
    portfolio_schema: dict, 
    unmatched_holdings: list, 
    batch_size: int, 
    start: int = 0, 
    previous_suggestions: list = None
):
    return """
You are a schema refinement assistant for financial tables. Your task is:
- Review a batch of {batch_size} unmatched financial holdings.
- Given the current schema (JSON below), propose new classes or modifications so each holding 
    can be classified.
- If a holding matches a previously suggested class, propose new optional fields if needed.
- Return your schema suggestions as a list of Pydantic SchemaSuggestion model objects.

Current Portfolio Schema:
{Portfolio.model_json_schema()}

Batch of unmatched holdings:
{unmatched_holdings[start : start + batch_size]}

Previously seen suggestions (optional, from prior batches):
{previous_suggestions if previous_suggestions else None}

For each unique holding, propose:
- A new schema class, or a modification to an existing class (add or refine fields).
- Specify all required and optional fields with Python type hints.
- If similar to an earlier suggestion, mark only new fields as optional.
- Provide a sample match (the original holding string).
- Output format: a Python list of SchemaSuggestion objects, as defined below.
"""


class SchemaSuggestion(BaseModel):
    name: str              # Name of new or modified schema class
    suggested_schema: str  # JSON schema for the instrument.
    example: str           # Example instrument seen in unmatched holdings
    
\end{lstlisting}
\caption{\textbf{Recommender Agent schema suggestion prompt, output model, and example LLM response.} The agent sees a batched portion of unmatched holdings to recommend new alterations to the Portfolio schema. This prompt is batch-specific and may include \texttt{previous\_suggestions} for cross-batch refinement and de-duplication.}
\label{app:scehma_recon}
\end{figure*}

\begin{figure*}[ht]
\begin{minipage}[t]{0.48\textwidth}
\begin{lstlisting}[language=json, caption={Currency Forward Generated JSON Schema}, ,numbers=left,
  numbersep=6pt,
  xleftmargin=2.2em, breaklines=true,
  breakatwhitespace=true,
  columns=fullflexible,
  keepspaces=true]
{
  "title": "Currency Forward",
  "type": "object",
  "properties": {
    "description": {
      "type": "string",
      "title": "Description",
      "description": "Description or name of the currency forward"
    },
    "market_value": {
      "anyOf": [
        { "type": "number" },
        { "type": "null" }
      ],
      "title": "Market Value",
      "description": "Market value of the currency forward",
      "default": null
    },
    "instrument_type": {
      "type": "string",
      "title": "Instrument Type",
      "const": "Currency Forward",
      "default": "Currency Forward"
    },
    "currency_pair": {
      "anyOf": [
        { "type": "string" },
        { "type": "null" }
      ],
      "title": "Currency Pair",
      "description": "Currency pair involved in the forward contract",
      "default": null
    },
    "forward_rate": {
      "anyOf": [
        { "type": "number" },
        { "type": "null" }
      ],
      "title": "Forward Rate",
      "description": "Agreed forward rate",
      "default": null
    },
    "settlement_date": {
      "anyOf": [
        { "type": "string", "format": "date-time" },
        { "type": "null" }
      ],
      "title": "Settlement Date",
      "description": "Settlement date for the currency forward",
      "default": null
    }
  }
}
\end{lstlisting}
\end{minipage}%
\hspace{0.05\textwidth}
\begin{minipage}[t]{0.46\textwidth}
\centering
  \begin{minipage}[t]{\textwidth}
    \begin{lstlisting}[language=Python, caption={Currency Forward Pydantic Model}, label={lst:cf-pydantic}]
class CurrencyForward(BaseModel):
    description: str
    market_value: Optional[float]
    instrument_type: str = "Currency Forward"
    currency_pair: Optional[str]
    forward_rate: Optional[float]
    settlement_date: Optional[datetime]
    \end{lstlisting}
  \end{minipage}
  \vspace{0.7em} 
  \begin{minipage}[t]{\textwidth}
    \begin{lstlisting}[language=json, caption={Refined Extraction}, label={lst:cf-example}]
# Raw inputs
"Bought EUR Sold USD at 0.93035372 11/06/2024"
"Bought USD Sold GBP at 1.25473636 31/05/2024"
"Bought GBP Sold USD at 0.79368122 16/05/2024"

# Extracted as fields
{
  "description": "Bought EUR Sold USD at 0.93035372 11/06/2024",
  "market_value": -282515.0,
  "instrument_type": "Currency Forward",
  "currency_pair": "EUR/USD",
  "forward_rate": 0.93035372,
  "settlement_date": "2024-06-11T00:00:00"
},
{
  "description": "Bought USD Sold GBP at 1.25473636 31/05/2024",
  "market_value": 20651.0,
  "instrument_type": "Currency Forward",
  "currency_pair": "USD/GBP",
  "forward_rate": 1.25473636,
  "settlement_date": "2024-05-31T00:00:00"
},
{
  "description": "Bought GBP Sold USD at 0.79368122 16/05/2024",
  "market_value": 1429313.0,
  "instrument_type": "Currency Forward",
  "currency_pair": "GBP/USD",
  "forward_rate": 0.79368122,
  "settlement_date": "2024-05-16T00:00:00"
}
    \end{lstlisting}
  \end{minipage}
\end{minipage}
\caption{\textbf{Left:} Final \texttt{Currency Forward} JSON schema. \textbf{Top right:} Equivalent Pydantic model. \textbf{Bottom right:} Example input string and its extraction into schema fields. This demonstrates schema-driven parsing of text into structured portfolio data. A currency forward contract is a financial instrument in the foreign exchange market that locks in the price at which an entity can buy or sell a currency at a future date.}
\label{app:example_recon}
\end{figure*}

\end{document}